\documentclass[10pt]{IEEEtran}
\usepackage{cite}
\usepackage{amsmath,amssymb}
 \usepackage{mathrsfs}
 \usepackage{mathtools}
\usepackage{dsfont}
\usepackage{tabularx}
\usepackage{graphicx}
\usepackage{epstopdf}
\usepackage{color}

\usepackage{algorithm}
\usepackage{algpseudocode}
\usepackage{algcompatible}

\usepackage{Tianpei_Report}

\begin{document}
\epstopdfsetup{outdir=./}
\allowdisplaybreaks
\setlength{\textfloatsep}{7pt}
\title{Robust training on approximated minimal-entropy set}

\author{Tianpei~Xie,
        Nasser~M.~Nasrabadi,~\IEEEmembership{Fellow,~IEEE,}
        and~Alfred~O.~Hero,~\IEEEmembership{Fellow,~IEEE}
\thanks{Tianpei Xie is with the Department
of Electrical and Computer Engineering, University of Michigan, Ann Arbor, 
MI., 48109, USA e-mail: (tianpei@umich.edu).}
\thanks{Nasser M. Nasrabadi is with the Lane Department of Computer
Science and Electrical Engineering, West Virginia University, Morgantown, WV, US. email: nasser.nasrabadi@mail.wvu.edu}
\thanks{Alfred O. Hero is with the Department
of Electrical and Computer Engineering, University of Michigan, Ann Arbor, 
MI., 48109, USA e-mail: (hero@umich.edu).}
\thanks{This research was supported by US Army Research Office (ARO) grants WA11NF-11-1-103A1.}}

\maketitle
\begin{abstract}

In this paper, we propose a general framework to learn a robust large-margin binary classifier when corrupt measurements, called anomalies, caused by sensor failure might be present in the training set. The goal is to minimize the generalization error of the classifier on non-corrupted measurements while controlling the  false alarm rate associated with anomalous samples. By incorporating a non-parametric regularizer based on an empirical entropy estimator,  we propose a Geometric-Entropy-Minimization regularized Maximum Entropy Discrimination (GEM-MED) method to learn to classify and detect anomalies  in a joint manner. We demonstrate  using simulated data and a real multimodal data set.  Our GEM-MED method can yield improved performance over  previous robust classification methods in terms of  both classification accuracy and anomaly detection rate.
\end{abstract}

\section{Introduction}\vspace{2pt}
Large margin classifiers, such as the support vector machine (SVM) \cite{scholkopf2002learning} and the maximum entropy discrimination (MED) classifier \cite{jaakkola1999maximum}, have enjoyed great popularity in the signal processing and machine learning communities due to their broad applicability, robust performance, and the availability of fast software implementations. When the training data is representative of the test data,  the performance of MED/SVM has theoretical guarantees that have been validated in practice \cite{bartlett2003rademacher, bousquet2002stability, scholkopf2002learning}. Moreover, since the decision boundary of the MED/SVM is solely defined by a few support vectors, the algorithm can tolerate random feature distortions and perturbations.

\begin{figure}[t] 
  \centering
  \centerline{\includegraphics[width=8.5cm]{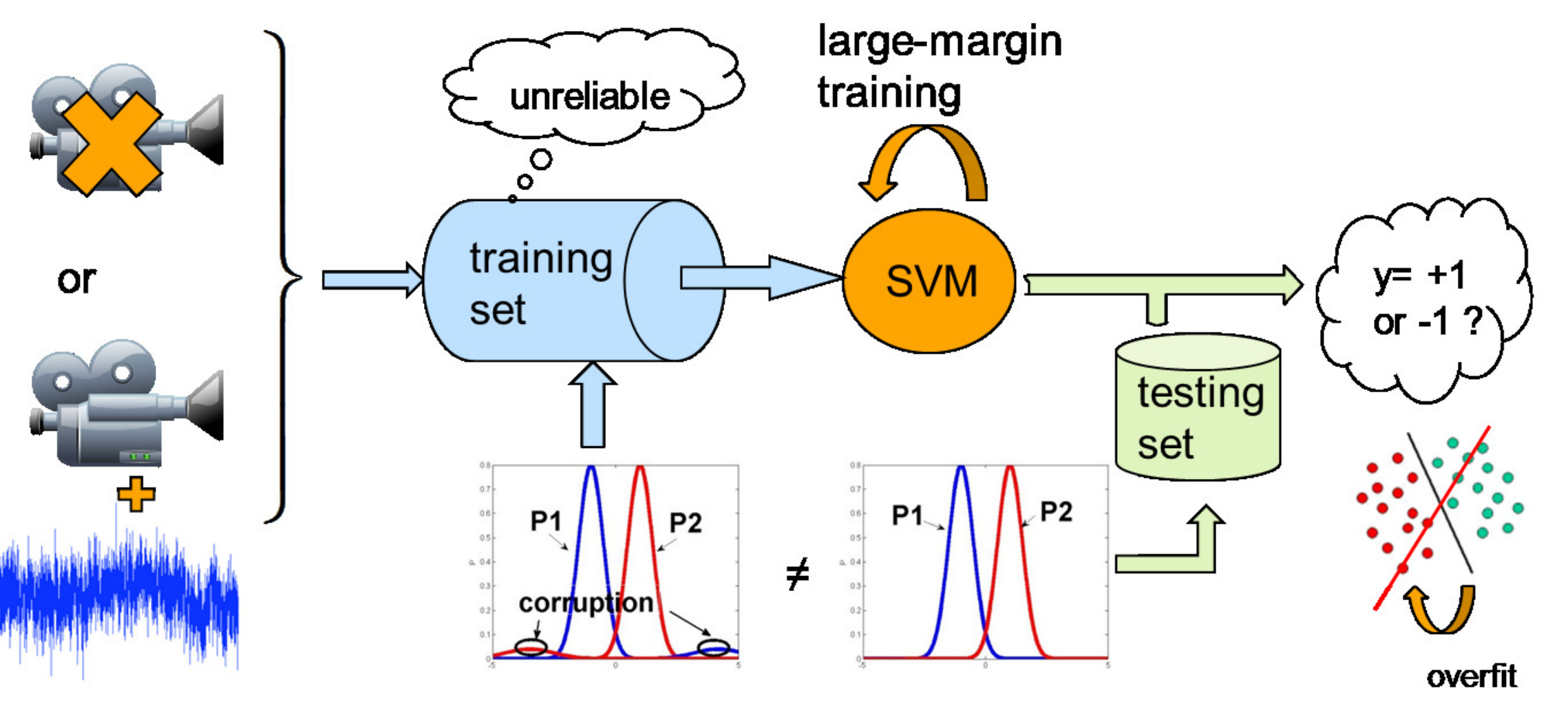}}
\caption{\footnotesize{ Due to  corruption in the training data  the training and testing sample  distributions are different from each other, which introduces errors into the decision boundary. }}
\label{fig: intro}
\end{figure}

However, in many real applications, anomalous measurements are inherent to the data set due to  strong environmental noise or possible sensor failures. Such anomalies arise in industrial process monitoring, video surveillance, tactical multi-modal sensing, robust spectrum sensing \cite{chen2008robust, ding2014robust}, and, more generally,  any application that involves unattended sensors in difficult environments (Fig. \ref{fig: intro}). Anomalous measurements are understood to be observations that have been corrupted, incorrectly measured, mis-recorded, drawn from different environments than those intended, or occurring too rarely to be useful in training a classifier \cite{yang2010relaxed}. If not robustified to anomalous measurements, classification algorithms may suffer from severe degradation of performance. Therefore, when anomalous samples are likely, it is crucial to incorporate outlier detection into the classifier design. 
This paper provides a new robust approach to design outlier resistant large margin classifiers. 

\vspace{-15pt}
  
\subsection{Problem setting and our contributions}  \vspace{-5pt}
We  divide the class of supervised training methods into four categories, according to how anomalies enter into different learning stages.\vspace{-10pt}  
\begin{table}[htb] 
\centering
\caption{\footnotesize Categories for supervised training algorithms via different assumption of anomalies} 
\label{tab: cate_anomalies}
\begin{tabular}{|m{50pt}<{\centering}|m{90pt}<{\centering}|m{85pt}<{\centering}|}
\hline 
 & Training set (uncorrupted) & Training set (corrupted) \\ 
\hline 
Test set (uncorrupted)  & classical learning algorithms (e.g. \cite{freund1995desicion, vapnik1998statistical, jaakkola1999maximum}) & Robust classification $\&$ training (e.g.  \cite{bartlett2003rademacher, tyler2008robust, bousquet2002stability, song2002robust, krause2004leveraging, xu2006robust, wu2007robust, wang2008training, masnadi2009design, long2010random, forero2012robust, ding2013kernel}, \textbf{this paper}) \\ 
\hline 
Test set (corrupted) & anomaly detection (e.g. \cite{scholkopf1999support, scott2006learning, hero2006geometric, sricharan2011efficient}) & Domain adaptation $\&$ transfer learning (e.g. \cite{blitzer2006domain, dai2007boosting, pan2010survey}) \\ 
\hline 
\end{tabular}
\end{table}\vspace{-10pt}

As shown in Table \ref{tab: cate_anomalies}, a  majority of learning algorithms assume that the training  and test samples  follow the same  nominal distribution and neither are corrupted by anomalies. Under this assumption, an empirical error minimization algorithm can achieve consistent performance on the 
test set.  In the case that anomalies exist only in the test data, one can apply  anomaly detection algorithms, e.g. \cite{chandola2009anomaly, scott2006learning, hero2006geometric, sricharan2011efficient}, to separate the anomalous samples from nominal ones. Under additional assumptions on the nominal set, these algorithms can effectively identify an anomalous sample under  given false alarm rate and miss rate. Furthermore, in the case that both training and test set are corrupted, possibly with different corruption rate, domain adaptation or transfer learning methods may be applied \cite{blitzer2006domain, daume2006domain, pan2010survey}.

This paper falls into the category of \emph{robust classification $\&$ training} in which  possibly anomalous samples occur in the training set. Such a problem is relevant, for example, when high quality clean training data is too expensive or too difficult to obtain. In \cite{bartlett2003rademacher, bousquet2002stability, krause2004leveraging}, the test set is assumed to be uncorrupted so that the test error provides an unbiased estimate of the generalization error on the \emph{nominal data set}, which is a standard measure of performance for robust classifiers. We adopt this assumption, although we also evaluate the proposed robust classifier when the test set is also corrupted with limited corruption rate.  
Our \emph{goal}  is to train a classifier that minimizes the generalization error with respect to the nominal data distribution when the \emph{training} set may be corrupted.  

The area of robust classification has been thoroughly investigated in both theory \cite{bartlett2003rademacher, tyler2008robust, bousquet2002stability, krause2004leveraging, xu2006robust, wu2007robust, masnadi2009design} and applications \cite{wang2008training, long2010random, forero2012robust, ding2013kernel}. Tractable robust classifiers that identify and remove outliers, called the Ramp-Loss based learning methods, have been studied in \cite{song2002robust,bartlett2003rademacher, xu2006robust,wang2008training}. Among these methods, Xu et al. \cite{xu2006robust} proposed the Robust-Outlier-Detection (ROD) method as an outlier detection and removal algorithm using the soft margin framework. Training the ROD algorithm involves solving an optimization problem, for which  dual solution is obtained via semi-definite programming (SDP). 
Like all the Ramp-Loss based learning models, this optimization is  non-convex requiring random restarts to ensure a globally optimal solution \cite{long2010random, yang2010relaxed}. 
In this paper, in contrast to the models above, a \emph{convex} framework for robust classification is proposed and a tractable algorithm is presented that finds the unique optimal solution  of a penalized entropy-based objective function. 
Our proposed algorithm is motivated by the basic principle underlying the so-called \emph{minimal volume (MV) /minimal entropy (ME) set anomaly detection method} \cite{scholkopf1999support, scott2006learning, hero2006geometric, sricharan2011efficient}. Such methods are expressly designed to detect anomalies in order to attain the lowest possible false alarm and miss probabilities. In machine learning, nonparametric algorithms are often preferred since they make fewer assumptions on the underlying distribution. 
Among these methods, we focus on  the Geometric Entropy Minimization (GEM) algorithm \cite{hero2006geometric, sricharan2011efficient}. This algorithm estimates the ME set based on the k-nearest neighbor graph (k-NNG), which is shown to be the Uniformly Most Powerful Test at given level when the anomalies are drawn from an unknown mixture of known nominal density and uniform anomalous density\cite{hero2006geometric}.  A \emph{key contribution} of this paper is the incorporation of the non-parametric GEM anomaly detection into a binary 
classifier under a non-parametric corrupt-data model. 


The proposed framework, called the \emph{Geometric-Entropy-Minimization regularized by Maximum Entropy Discrimination (GEM-MED)}, 
follows a \emph{Bayesian} perspective. It is an extension of the well-established Maximum Entropy Discrimination (MED) approach proposed by Jaakkola et al. \cite{jaakkola1999maximum}. MED performs Bayesian large margin classification via the maximum entropy principle and it subsumes SVM as a special case. The MED model can also solve the parametric anomaly detection \cite{jaakkola1999maximum} problem and has been extended to multitask classification \cite{jebara2011multitask}. A naive application of MED to robust classification might use a two-stage approach that implements an anomaly detector on the training set prior to training the MED classifier, which is sub-optimal. In this paper, we propose GEM-MED as a unified approach that jointly solves an anomaly detection  and classification problem via the MED framework. The GEM-MED explicitly incorporates the anomaly detection false-alarm constraint and the mis-classification rate constraint into a maximum entropy learning framework. Unlike the two-stage approach, GEM-MED finds anomalies by investigating both the underlying sample distribution and the sample-label relationship, allowing anomalies in support vectors to be more effectively suppressed. As a Bayesian approach, GEM-MED requires no tuning parameter as compared to other anomaly-resistant classification approaches, such as ROD \cite{xu2006robust}. 
We demonstrate the superior performance of the GEM-MED anomaly-resistant classification approach over other robust learning methods on simulated data and on a real data set combining sensor failure. The real data set contains human-alone and human-leading-animal footsteps, collected in the field by an acoustic sensor array \cite{damarla2011detection, damarla2012seismic, huang2011multi}. 
\vspace{-15pt}
\subsection{Organization of the paper}\vspace{-10pt}
What follows is a brief outline of the paper. In Section \textbf{II}, we review  MED as a general framework to perform classification and other inference tasks.  The proposed combined GEM-MED approach is presented in Section \textbf{III}.  A variational implementation of GEM-MED  is introduced in Section \textbf{IV}.  Experimental results based on synthetic data and real data are presented in Section \textbf{V}. Our conclusions are discussed in Section \textbf{\ref{lab: conclude}}. 

\vspace{-12pt}
\section{From MED to GEM-MED: a general routine}\vspace{-5pt}
Denote the training data set as $\mathcal{D}_{t} \defeq \{  (y_{n}, \mathbf{x}_{n} )\}_{n\in T}$,  where each sample-pair $(y_{n}, \mb{x}_{n} )\in \cY \times \cX = \cD$ are independent.  Denote the feature set $\cX \subset \mathcal{R}^{p}$ and the label set as $\cY$. For simplicity, let $\cY= \{-1,1\}$. The test data set is denoted as $\mathcal{D}_{s}\defeq \{ \mathbf{x}_{m}\}_{m\in S}$. 
We assume that $\{  (y_{n}, \mathbf{x}_{n} )\}_{n\in T}$ are i.i.d. realizations of random variable $(Y,X)$ with distribution $\cP_{t}$, conditional probability density $p(X| Y=y, \Theta)$ and prior $p(Y=y), y\in \cY$, where $\Theta$ is the set of unknown model parameters.  We denote by $p(Y=y, X; \Theta) = p(X| Y=y, \Theta)p(Y=y)$ the parameterized joint distribution of $(Y,X)$. The distribution of test data, denoted as $\cP_{s}$, is defined similarly. $\cP_{nom}$ denotes  the nominal distribution. Finally, we define the probability simplex $\Delta_{\cY\times \cX}$ over the space $\cY\times \cX$.
\vspace{-10pt}
\subsection{MED for classification and parametric anomaly detection} \label{sec 2_1}\vspace{-5pt}
The Maximum entropy discrimination (MED) approach to learning a classifier was proposed by Jaakkola et al \cite{jaakkola1999maximum}. The MED approach is a Bayesian maximum entropy learning framework that can either perform conventional classification, when $\cP_{t} = \cP_{s}= \cP_{nom}$, or  anomaly detection, when $\cP_{t}\neq \cP_{s}$, and $\cP_{t} = \cP_{nom}$. In particular, assume that all parameters in $\Theta$ are random with prior distribution $p_{0}(\Theta)$. Then MED is formulated as finding the posterior distribution $q(\Theta)$ that minimizes the relative entropy  
\begin{align}
\kl{q(\Theta)}{p_{0}(\Theta)} &\defeq \int \log\paren{\frac{q(\Theta)}{p_{0}(\Theta)}} q(d\Theta) \label{expr: MED}
\end{align} 
subject to a set of $P$ constraints on the risk or loss: 
\begin{align}
\phantom{=} \int \cL_{i}\paren{p, (y_{n}, \mb{x}_{n}); \Theta}q(d\Theta) \le 0,\; \forall n \in T, 1\le i\le P. \label{expr: error_generative}
\end{align}
The constraint functions $\set{\cL_{i}}_{i=1}^{P}$ can correspond to losses associated with different type of errors, e.g. misdetection, false alarm  or misclassification. For example, the classification task  defines a parametric discriminant function $\cF_{C}:  \Delta_{\cY\times \cX} \times \cD \rightarrow \bR_{+}$ as 
\begin{align*}
\cF_{C}\paren{p, (y_{n}, \mb{x}_{n}); \Theta} &\defeq \log p(Y=y_{n}| \mb{x}_{n}; \Theta)/p(Y\neq y_{n}| \mb{x}_{n}; \Theta).
\end{align*} In the case of the SVM classification, the loss function is defined as 
\begin{align}
\cL_{i} = \cL_{C}\paren{p, (y_{n}, \mb{x}_{n}); \Theta} &\defeq \brac{\xi_{n}-  \cF_{C}\paren{p, (y_{n}, \mb{x}_{n}); \Theta} }. \label{expr: error_hinge}
\end{align} Other definitions of discriminant functions are also possible \cite{jaakkola1999maximum}.

An example of an anomaly detection test function $\cL_{i} = \cL_{D}: \Delta_{\cX} \times  \cX  \rightarrow \bR$, is
\begin{align}
\cL_{D}\paren{p, \mb{x}_{n}; \Theta}&\defeq -\brac{\log p(\mb{x}_{n}; \Theta) - \beta},  \label{expr: anom_test}
\end{align}
where $p(\mb{x}_{n}; \Theta)$ is the marginal likelihood $p(\mb{x}_{n}; \Theta)= \sum_{y_{n}\in \cY}p(X=\mb{x}_{n}|Y=y_{n},\Theta)p(Y=y_{n})$. The constraint function \eqref{expr: anom_test} has the interpretation as local entropy of $X$ in the neighborhood of $X=\mb{x}_{n}$. Minimization of the \emph{average} constraint function yields the minimal entropy anomaly detector  \cite{hero2006geometric, sricharan2011efficient}.
The solution to the minimization \eqref{expr: error_generative} yields a posterior distribution $p(Y=y|\mb{x}_{n}, \overline{\Theta})$ where $\overline{\Theta} \defeq \Theta\cup \set{\xi_{n}}\cup\set{\beta}$.  This lead to a discrimination rule
\begin{align}
y^{*} = \argmin_{y} \set{ -\int \log p(y, \mb{x}_{m}\,;\, \overline{\Theta}) q(d\overline{\Theta})   }, \, \mb{x}_{m}\in \cD_{s}. \label{expr: test_label_MED}
\end{align} when applied to the test data $\cD_{s}$. 

The decision region $\set{\mb{x}\in \cX| Y= y}$  of MED can have various interpretations  depending on the form of the constraint function \eqref{expr: error_hinge} and \eqref{expr: anom_test}. For the anomaly detection constraint \eqref{expr: anom_test}, it is easily seen that the decision region is a \emph{$\beta$-level-set region} for the marginal  $p(\mb{x}; \overline{\Theta})$, denoted as $\Psi_{\beta} \defeq $ $\{\mb{x}_{n}\in \cX \,|\, $ $ \log p(\mb{x}_{n}; \overline{\Theta}) $ $\ge \beta  \}$. Here $\Psi_{\beta}$ is the \emph{rejection region} associated with the test: declare $\mb{x}_{m}\in \cD_{s}$ as anomalous whenever $\mb{x}_{m} \not\in \Psi_{\beta}$; and declare it as nominal if $\mb{x}_{m} \in \Psi_{\beta}$. 
With a properly-constructed decision region, the MED model, as a projection of prior distribution $p_{0}(\Theta)$ into this region, can  provide  performance guarantees in terms of the error rate or the false alarm rate  and can result in improved accuracy \cite{jebara2011multitask, zhu2011infinite}.




Similar to the SVM, the MED model  readily  handles nonparametric classifiers. For example, the discriminant function $\cF_{C}\paren{p, (y, \mb{x}); \Theta} $ can take the form $y[\Theta(\mb{x})]$ where $\Theta=f$ is a random function, and $f: \cX \rightarrow \cY$ can be specified by a Gaussian process with Gaussian covariance kernel $K(\cdot , \cdot).$ More specifically,  $f \in \cH$, where $\cH$ is a Reproducing Kernel Hilbert Space (RKHS) associated with kernel function $K: \cX \times \cX \rightarrow \bR$. See \cite{jebara2011multitask} for more detailed discussion.

MED utilizes a  weighted ensemble strategy that can improve the classifier stability \cite{jaakkola1999maximum}. However, like SVM, MED is sensitive to anomalies in the training set. 
\vspace{-15pt} 

\subsection{Robustified MED when there is an anomaly detection oracle}\vspace{-10pt}
Assume an \emph{oracle} exists that identifies anomalies in the training set. Using this oracle, partition the training set as $\cD_{t} = \cD_{t}^{nom}\cup \cD_{t}^{anm}$, where $(\mb{x}_{n},y_{n})\sim \cP_{nom}$ if $(\mb{x}_{n},y_{n})\in \cD_{t}^{nom}$ and $(\mb{x}_{n},y_{n})\not\sim \cP_{nom}$, if $(\mb{x}_{n},y_{n})\in \cD_{t}^{anm}$. Given the oracle, one can achieve robust classification simply by constructing a classifier and an anomaly detector simultaneously on $\cD_{t}^{nom}$. 
This results in a naive implementation of robustified MED  as \vspace{-2pt}
\begin{align}
\min_{q(\overline{\Theta})\in \Delta_{\overline{\Theta}}} & \phantom{=} \kl{q(\overline{\Theta})}{p_{0}(\overline{\Theta})} \label{expr: robust_MED}\\
\text{s.t. }&\int \cL_{C}\paren{p, (y_{n}, \mb{x}_{n}); \overline{\Theta}}q(d\overline{\Theta})\le 0, \, (\mb{x}_{n},y_{n})\in \cD_{t}^{nom}, \label{expr: robust_MED_ineq1}\\
&\int \cL_{D}\paren{p, \mb{x}_{n}; \overline{\Theta}}q(d\overline{\Theta}) \le 0,\;(\mb{x}_{n},y_{n})\in \cD_{t}^{nom}, \label{expr: robust_MED_ineq2}
\end{align}
where $\overline{\Theta}  = \Theta\cup \set{\beta}\cup \set{\xi_{n}}_{n\in T}$, the large-margin error function $\cL_{C}$  is defined in \eqref{expr: error_hinge}  and the test function $\cL_{D}$ is defined in \eqref{expr: anom_test}.
The prior is defined as $p_{0}(\overline{\Theta})= p_{0}(\Theta)p_{0}(\beta)\prod_{n\in T}p_{0}(\xi_{n})$. 

Of course, the oracle partition $\cD_{t} = \cD_{t}^{nom}\cup \cD_{t}^{anm}$ is not available \emph{a priori}. The parametric estimator $\widehat{\Psi}_{\beta}$ of $\Psi_{\beta}$ can be introduced in place of $\cD_{t}^{nom}$ in \eqref{expr: robust_MED}. However, the estimator $\widehat{\Psi}_{\beta}$ is difficult to implement and can be severely biased if there is model mismatch. 
Below, we propose an alternative nonparametric estimate of the decision region $\Psi_{\beta}$ that learns the oracle partition. 
  
\vspace{-10pt}
 \section{The GEM-MED:  model formulation }
 \subsection{Anomaly detection using minimal-entropy set}
\begin{figure}[t] 
  \centering
  \centerline{\includegraphics[width=6.6cm]{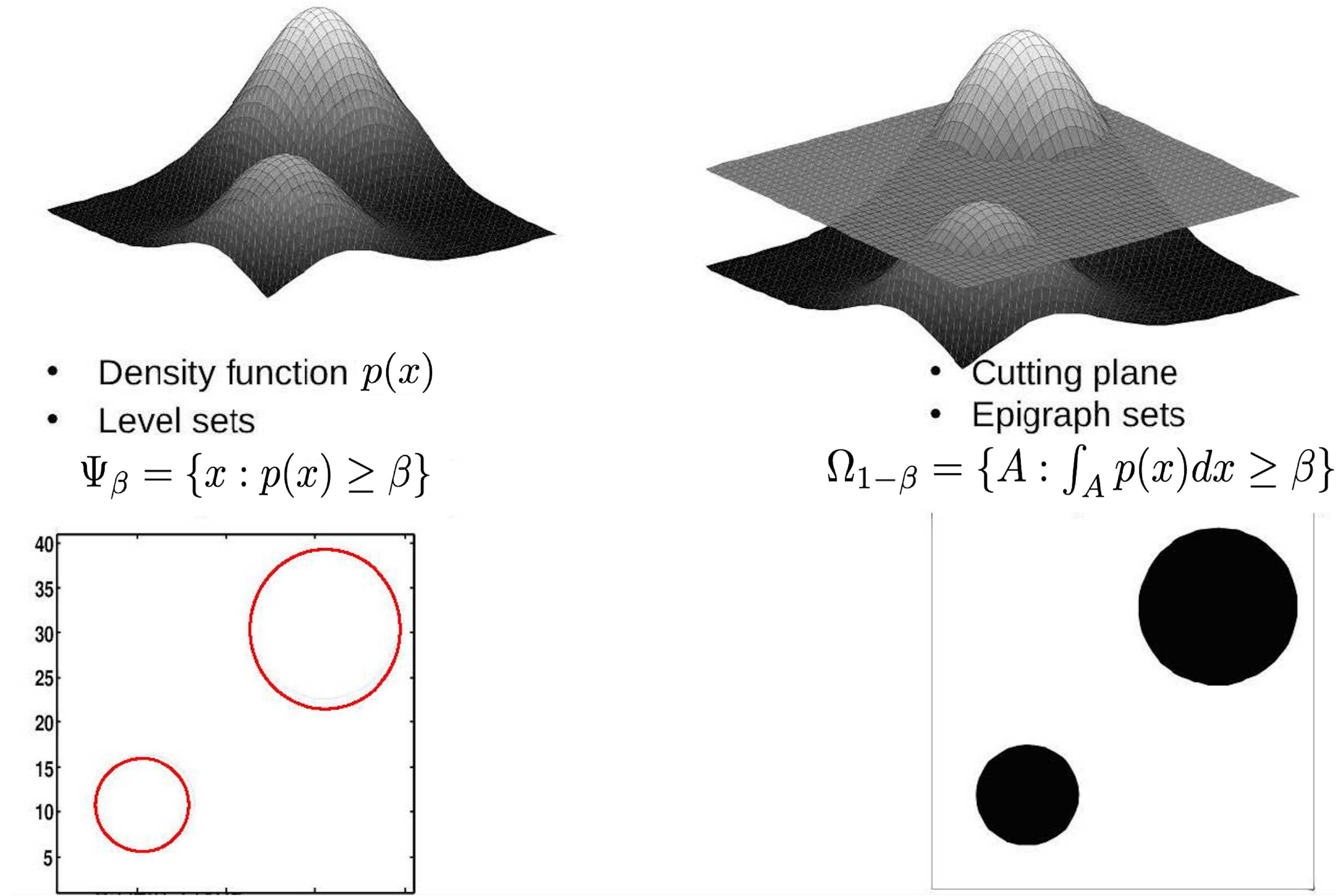}}
\caption{\footnotesize{ The comparison of level-set (left) and the  epigraph-set (right) w.r.t. two continuous density function $p(x)$. The minimum-entropy-set is computed based on the epigraph-set. }}
\label{fig: epigraph-set}
\end{figure}

\vspace{-10pt}
As an alternative to a  parametric estimator of the level-set $\Psi_{\beta} \defeq \{\mb{x}_{m}\in \cX\,|\, \log p(\mb{x}_{m}; \overline{\Theta}) \ge \beta  \}$, we propose to use a non-parametric estimator \cite{wasserman2010all} based on the\emph{ minimal-entropy (ME) set} $\Omega_{1-\beta}$. The ME set $\Omega_{1-\beta} \defeq \arg\min_{A}\{H(A)|\,\,\hspace{-5pt}\int_{A}p(\mathbf{x})d{\mathbf{x}} \ge  \beta \} $ is referred as the \emph{minimal-entropy-set of false alarm level $1-\beta$}, where $H(A) = -\int_{A}\log p(\mathbf{x})\,p(\mathbf{x})d{\mathbf{x}} $ is the Shannon entropy of the density $p(\mathbf{x})$ over the region $A$.  This minimal-entropy-set is equivalent to the  \emph{epigraph-set} $\{A: \; \int_{A}p(\mathbf{x})d{\mathbf{x}} \ge \beta\}$ as illustrated in Fig. \ref{fig: epigraph-set}.



Given $\Omega_{1-\beta}$, the ME anomaly test is as follows: a sample $\mathbf{x}_{n}$ is declared  anomalous if $\mathbf{x}_{n} \not\in \Omega_{1-\beta}$; and it is declared nominal, when $\mathbf{x}_{n} \in \Omega_{1-\beta}$. It is established in \cite{hero2006geometric} that when $p(x)$ is a known density, this test is a Uniformly Most Powerful Test (UMPT) \cite{scharf1991statistical} at level $\beta$ of the hypothesis $H_{0}: x\sim p(x)$ vs. $H_{1}: x\sim p(x)+ \epsilon U(x)$, where $U(x)$ is the  uniform density and $\epsilon\in [0,1]$ is an unknown mixture coefficient.




\vspace{-15pt}

\subsection{The BP-kNNG implementation of GEM}\label{sec: GEM} 
\begin{figure}[tb] 
  \centering
  \begin{minipage}[b]{0.8\linewidth}
  \centering
  \centerline{\includegraphics[width=8.2cm, clip=true, trim= 1mm 80mm 10mm 20mm]{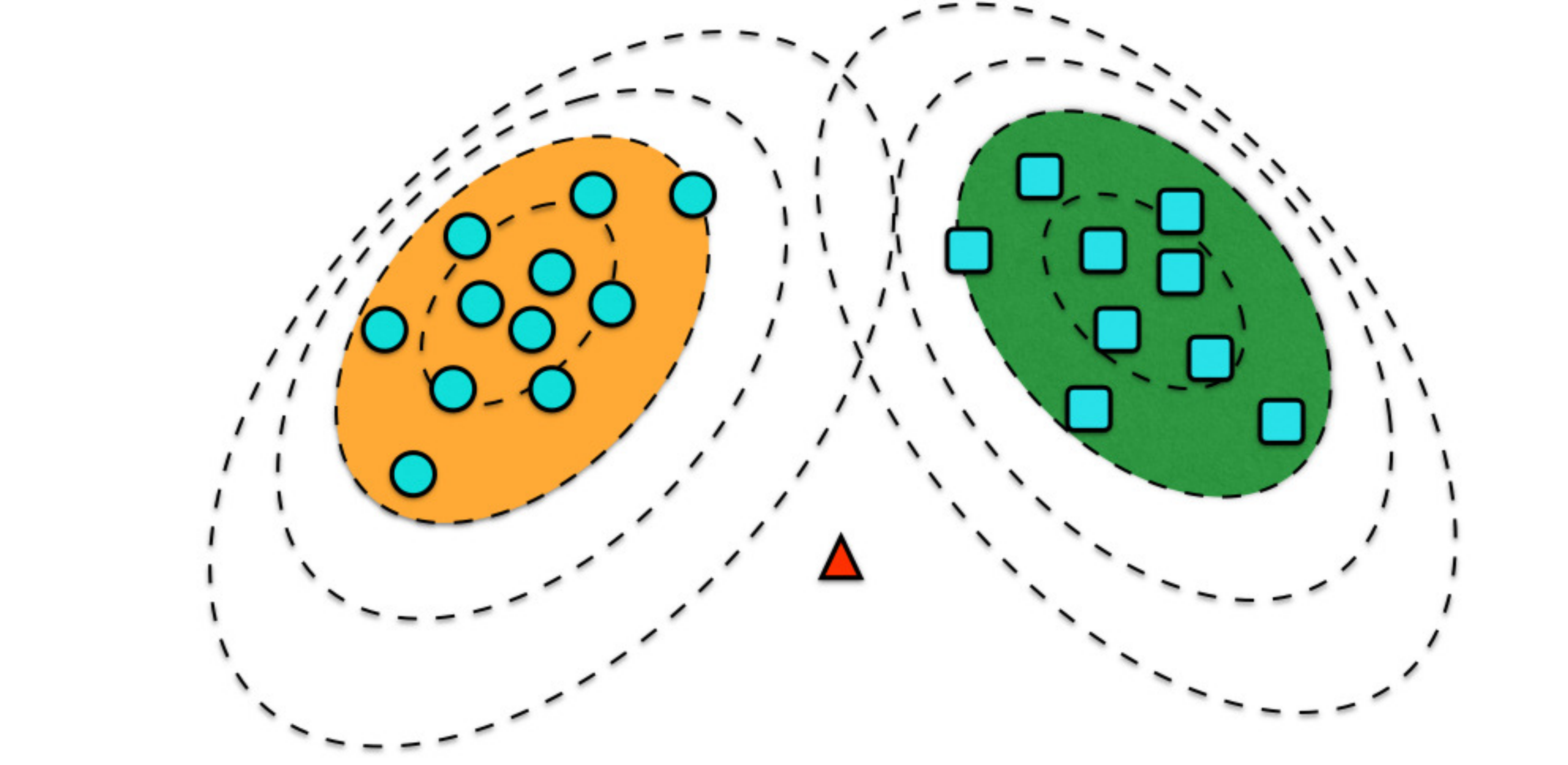}}
  \vspace{-7pt}
  \centerline{\footnotesize{(a)}}
\end{minipage}\\
\begin{minipage}[b]{0.6\linewidth}
  \centering
  \centerline{\includegraphics[width=8.2cm, clip=true, trim= 1mm 80mm 10mm 20mm]{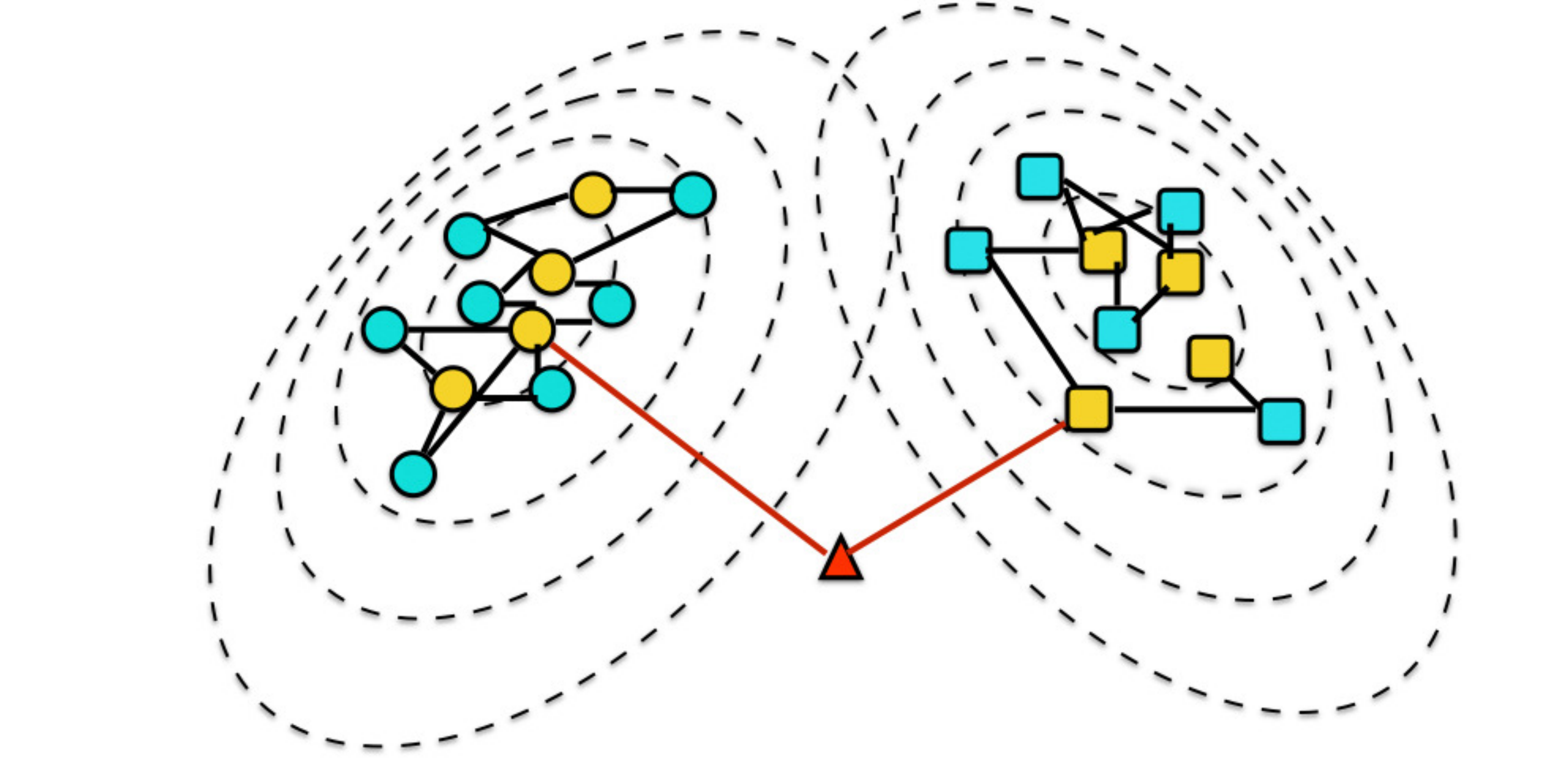}}
  \vspace{-10pt}
  \centerline{\footnotesize{(b)}}
\end{minipage}\vspace{-7pt}
\caption{\footnotesize{ Figure (a) illustrates ellipsoidal minimum entropy (ME) sets for two dimensional Gaussian features in the training set for class 1 (orange region) and class 2 (green region). These ME sets have coverage probabilities $1-\beta$ under each class distribution and correspond to the regions of maximal concentration of the densities. The blue disks and blue squares inside these regions correspond to the nominal training samples under class 1 and class 2, respectively. An outlier (in red triangle) falls outside of both of these regions.  Figure (b) illustrates the bipartite 2-NN graph approach to identify the anomalous point, where the yellow disks and squares are reference samples  in each class that are  randomly selected from the training set. Note that  the average 2-NN distance for anomalies should be significantly larger than that for the nominal samples.}}   
\label{fig: BP-KNNG}
\end{figure}

\vspace{-10pt} 
Several methods have been proposed to empirically approximate the ME set $\Omega_{1-\beta}$ including: kernel density estimation \cite{scott2006learning}; the $k$-point minimal spanning tree \cite{hero1999asymptotic}; the leave-one-out $k$-nearest-neighbor graph \cite{sricharan2011efficient}; and the average $k$-nearest-neighbor distance \cite{root2015learning}.  In 
\cite{sricharan2011efficient}, the bipartite k-nearest-neighbor (BP-kNN) based algorithm was proposed as an alternative approximation.  The BP-kNN solves the following discrete optimization problem: 
\begin{align*}
A^{*}_{c}\in \arg\min_{A_{c}\subset  \cD_{t}^{N,c}} L(A_{c}, \cD_{t}^{M,c}), \\
\text{where } L(A_{c}, \cD_{t}^{M,c}) \defeq \sum_{\mb{x}_{n}\in A_{c}}d_{k}(\mb{x}_{n}, \cD_{t}^{M,c}),  
\end{align*} and where $A_{c}$ is a set of distinct $K=\abs{T}(1-\beta)$ points in $\cD_{t}^{N,c}$  (see Fig. \ref{fig: BP-KNNG} for illustration). It is shown in \cite{sricharan2011efficient} that $A^{*}_{c}= \widehat{\Omega}_{1-\beta}$ is an asymptotically consistent estimator of the ME set. Equivalently, let $\eta_{n}\in \set{0,1}$ be the indicator function of the event $\mathbf{x}_{n} \in A_{c}$ and define $d_{n} \defeq d_{k}(\mb{x}_{n}, \cD_{t}^{M,c})$.  Then it can easily be shown that  the algorithm in  \cite{sricharan2011efficient}  finds the optimal  binary variables $\set{\eta_{n}\in \set{0,1}| \mb{x}_n\in \cD_{t}^{N,c}}, n=1,\ldots, N,$ that minimize
\begin{align}
\sum_{\mb{x}_{n}\in  \cD_{t}^{N,c}}\eta_{n}d_{n} \quad \text{subject to }\sum_{\mb{x}_{n}\in  \cD_{t}^{N,c}}\eta_{n}\ge  K. \label{eqn: BP-kNNG_obj_weight}
\end{align}  
This representation makes the BP-kNN implementation of GEM naturally adaptable to our GEM-MED framework. Specifically,  the binary weights $\eta_n \in \{0,1\}$ are relaxed to continuous weights in the unit interval $[0,1]$ for all $n\in T$. After relaxation, the constraint in \eqref{eqn: BP-kNNG_obj_weight} becomes $\sum_{n}\eta_{n}/\abs{T}\ge  \hat{\beta}$, where $\hat{\beta}= K/\abs{T}=(1-\beta) >0$ is set so that the optimal solution $\set{\eta_{n}| \mb{x}_{n}\in A^{*}_{c}}$ is feasible and the all-zero solution is infeasible.
With the set of weights $\set{\eta_{n}}_{n\in T}$, the GEM problem in \eqref{eqn: BP-kNNG_obj_weight} can be transformed into a  set of nonparametric constraints that fit the framework  \eqref{expr: robust_MED}. This is discussed below.\vspace{-15pt}

\subsection{The GEM-MED as non-parametric robustified MED}\label{sec: GEM-MED}\vspace{-5pt}
Now we can implement the framework in \eqref{expr: robust_MED}.  
Denote $\overline{\Theta}\defeq \Theta\cup\set{\hat{\beta}}\cup \set{\xi_{n}}_{n\in T}\cup \set{\eta_{n}}_{n\in T} \cup \set{\gamma_{z}}_{z\in\set{\pm 1}}$, where $\Theta, \set{\xi_{n}}_{n\in T}$ are parameters as defined in \eqref{expr: robust_MED}, $\set{\eta_{n}}_{n\in T}$ are weights in Sec. \ref{sec: GEM} and $\hat{\beta}, \set{\gamma_{z}}_{z\in\set{\pm 1}}$ are variables to be defined later. \vspace{-5pt}
 
According to the objective function in \eqref{eqn: BP-kNNG_obj_weight}, we specify the test function $\tilde{\cL}_{D}$ as
\begin{align*}
&\tilde{\cL}_{D}(\overline{\Theta}, \mb{y}; z, \mb{d}) \defeq  \tilde{\cL}_{D}(\set{\eta_{n}}, \set{\gamma_{z}}, \mb{y}; z, \mb{d})\\
&= \paren{ \sum_{n}\ind{y_{n}= z}\eta_{n}d_{n}/\abs{T} - \gamma_{z}}, \quad z\in \set{\pm 1},
\end{align*} where $\gamma_{z}\ge 0, z\in \set{\pm 1}$ is the threshold associated with $d_{n}$ on $\cD_{t}\cap \set{\mb{x}_{n}| y_{n}=z}$. Compared with \eqref{eqn: BP-kNNG_obj_weight}, if  $\gamma_{z}= L_{z}^{*}+\epsilon$, where $L_{z}^{*}$ is the optimal value in \eqref{eqn: BP-kNNG_obj_weight} and  $\epsilon>0$ is small enough, then for $\set{\eta_{n}}_{n\in T}$ satisfying $\tilde{\cL}_{D}\le 0$, the region  $\set{\mb{x}_{n}: \eta_{n}>\frac{1}{2}}$ is concentrated on $\widehat{\Omega}_{1-\beta}\cap \set{\mb{x}_{n}| y_{n}= z}, z\in \set{\pm 1}$. 



\vspace{-5pt}

As discussed in \ref{sec: GEM}, the constraint in  \eqref{eqn: BP-kNNG_obj_weight} becomes the inequality constraint $ \sum_{n| y_{n}=z}\eta_{n}/\abs{T} \ge \hat{\beta}$.  


Assuming that $\overline{\Theta}$ is random with unknown distribution $q(\overline{\Theta})$, the above expected constraints becomes  \begin{align}
&\int \tilde{\cL}_{D}(\overline{\Theta}, \mb{y}; z, \mb{d})q(d\overline{\Theta}) \le 0, z\in \set{\pm 1}, \label{expr: entropy_constraint}\\
&\int \brac{\sum_{n: y_{n}=z}\eta_{n}/\abs{T}}q(d\overline{\Theta}) \ge \hat{\beta}, z\in \set{\pm 1}. \label{expr: epigraph_constraint}
\end{align}
The constraint \eqref{expr: entropy_constraint} is referred as \emph{the entropy constraint} and constraint \eqref{expr: epigraph_constraint} is the \emph{epigraph constraint}. As discussed above, the region $\set{\mb{x}_{n}| \eta_{n}>\frac{1}{2}}$ for $q(\overline{\Theta})$ satisfying \eqref{expr: entropy_constraint} and \eqref{expr: epigraph_constraint} is concentrated on $\widehat{\Omega}_{1-\beta}\cap \set{\mb{x}_{n}| y_{n}= z}$ in each class $z\in \set{\pm 1}$ on average. 
With $\tilde{\cL}_{D}$,  the test constraint
in \eqref{expr: robust_MED} is replaced by \eqref{expr: entropy_constraint} and \eqref{expr: epigraph_constraint}.

For the classification part in \eqref{expr: robust_MED}, given $\eta_{n}$ associated with each sample, the error constraints
in \eqref{expr: robust_MED} is replaced by \emph{reweighted} error constraints
\begin{align*}
\int \brac{\eta_{n}\cL_{C}\paren{p, (y_{n}, \mb{x}_{n}); \overline{\Theta}}}q(d\overline{\Theta})\le 0, \;n\in T,
\end{align*} with $\cL_{C}$ defined as in \eqref{expr: error_hinge}. Note that these constraints are applied to the entire training set. Summarizing, we have the following:\vspace{-10pt}
\begin{definition}
The \emph{Geometric-Entropy-Minimization Maximum-Entropy-Discrimination (GEM-MED)} method solves
\begin{align}
\min_{q(\overline{\Theta})\in \Delta_{\overline{\Theta}}} &\phantom{=} \kl{q(\overline{\Theta})}{p_{0}(\overline{\Theta})} \label{expr: GEM-MED}\\
\text{s.t. }&  \int \brac{\eta_{n}\cL_{C}\paren{p, (y_{n}, \mb{x}_{n}); \overline{\Theta}}}q(d\overline{\Theta})\le 0, \;n\in T,  \nonumber\\
&\int \tilde{\cL}_{D}(\overline{\Theta}, \mb{y}; z, \mb{d})q(d\overline{\Theta}) \le 0, \, z\in \set{\pm 1}, \nonumber\\
&\int \brac{\sum_{n: y_{n}=z}\eta_{n}/\abs{T}}q(d\overline{\Theta}) \ge \hat{\beta},\, z\in \set{\pm 1} \nonumber
\end{align}
where $\overline{\Theta}$, $\cL_{C}$ and $\tilde{\cL}_{D}$ are defined as before. 
\end{definition}

\vspace{-10pt}

\vspace{-5pt}
\section{Implementation}\label{sec: Implement}\vspace{-5pt}
\subsection{Projected stochastic gradient descent algorithm}\vspace{-5pt}
Note that \eqref{expr: GEM-MED} is a convex optimization w.r.t. the unknown distribution $q(\overline{\Theta})$ \cite{jaakkola1999maximum, cover2012elements}. Therefore, it can be solved using the Karush-Kuhn-Tucker (KKT) conditions, which will result in a unique solution.  We make the following simplifying assumptions under which our a computational algorithm is derived to solve \eqref{expr: GEM-MED}. \vspace{-8pt}
\begin{enumerate}
\item Assume that a kernelized SVM is used for the classifier discriminant $\cF_C$ function. Following \cite{zhu2014bayesian, jebara2011multitask}, we assume that the decision function $f$ follows a Gaussian random process on $\cX$, i.e., a positive definite covariance kernel  $K(\mb{x}_{i}, \mb{x}_{j})$ is defined for all $\mb{x}_i,\mb{x}_j \in \cX$ and all finite dimensional distributions, i.e., distributions of samples $(f(\mb{x}_{i}) )_{i\in T}, $ follow the multivariate normal distribution
\begin{align}
(f(\mb{x}_{i}))_{ i\in T}&\sim  \cN(\mb{0}, \mb{K}), \label{expr: Gaussian_prior}
\end{align}
where $\mb{K} = [ K(\mb{x}_{i}, \mb{x}_{j})]_{i,j\in T}$  is a specified covariance matrix. For example, $ K(\mb{x}_{i}, \mb{x}_{j}) \defeq \exp(-\gamma\| \mathbf{x}_{i} - \mathbf{x}_{j}\|_{2}^{2}) $ for Gaussian RBF kernel covariance function. 

\item Assume a separable prior, as commonly used in Bayesian inference \cite{jaakkola1999maximum, blei2003latent,  zhu2014bayesian}
\begin{align}
p_{0}(\overline{\Theta}) =p_{0}(\Theta)\prod_{n\in T}p_{0}(\xi_{n})\prod_{n\in T}p_{0}(\eta_{n})\prod_{z\in \set{\pm 1}}p_{0}(\gamma_{z}).  \label{expr: prior_factor}
\end{align}

\item Assume that the hyperparameters $\set{\xi_{n}}$ are exponential random variables and the indicator variables $\set{\eta_{n}}$ are independent Bernoulli random variables, 
\begin{align}
p_{0}(\xi_{n})&\propto \exp(-c_{\xi}(1- \xi_{n})),\; \xi_{n}\in (-\infty,1], \; n\in T; \nonumber\\
p_{0}(\eta_{n})&= Ber(p_{\eta})\nonumber\\
& \text{ with }p_{\eta}= \frac{1}{1+ \exp(-(a_{\eta}- \eta_{n}))}\nonumber\\
&\phantom{===}\defeq \sigma(a_{\eta}- \eta_{n}),\;\eta_{n}\in \set{0,1}, \; n\in T; \nonumber\\
p_{0}(\gamma_{z}) &= \delta_{\hat{\gamma}_{z}}(\gamma_{z}); \; z\in \set{\pm 1}, \label{expr: Other_prior}
\end{align}
where $(a_{\eta}, c_{\xi})$ are parameters and $\hat{\gamma}_{z}$ is the upper bound estimate for minimal-entropy in each class $z=\pm 1$ given by GEM algorithm. $\sigma(x) = 1/(1+ \exp(-x))$ is the sigmoid function.
\end{enumerate}

Now by solving  the primal version of optimization problem \eqref{expr: GEM-MED},  we have 
\begin{theorem}\label{prop: unique_solution}
The GEM-MED problem in \eqref{expr: GEM-MED} is convex with respect to the unknown distribution $q(\overline{\Theta})$ and the unique optimal solution is a \emph{generalized Gibbs distribution} with the density:
\begin{align}
q(d\overline{\Theta})&= \frac{1}{Z(\mb{\lambda}, \mb{\mu}, \mb{\kappa})}p_{0}(d\overline{\Theta})\exp\paren{-E(\overline{\Theta}; \mb{\lambda}, \mb{\mu}, \mb{\kappa})}, \label{expr: opt_general}
\end{align}
where 
\begin{align*}
E(\overline{\Theta}; \mb{\lambda}, \mb{\mu}, \mb{\kappa}) &\defeq E(\Theta, \hat{\beta}, \set{\xi_{n}}, \set{\eta_{n}}, \set{\gamma_{z}} ; \mb{\lambda}, \mb{\mu}, \mb{\kappa})\\
&= \sum_{n\in T}\lambda_{n}\eta_{n}\cL_{C,\Theta,\xi_{n}}- \sum_{z\in \set{\pm 1}}\mu_{z}\tilde{\cL}_{D,z}\\
&\phantom{=}- \sum_{z\in \set{\pm 1}}\kappa_{z}\sum_{n: y_{n}=z}\eta_{n}/\abs{T}+ \sum_{z\in \set{\pm 1}}\kappa_{z}\hat{\beta}\nonumber
\end{align*} 
with $\overline{\Theta}= \Theta\cup\set{\hat{\beta}}\cup \set{\xi_{n}}_{n\in T}\cup \set{\eta_{n}}_{n\in T} \cup \set{\gamma_{+1}, \gamma_{-1}}$and where the dual variables $\mb{\lambda}=\set{\lambda_{n}, n\in T}$, $\mb{\mu}= (\mu_{z}, z\in \pm 1)$ and $\mb{\kappa}= (\kappa_{z}, z\in \pm 1)$ are all nonnegative. $Z(\mb{\lambda}, \mb{\mu}, \mb{\kappa})$ is \emph{the partition function}, which is given as
\begin{align}
Z(\mb{\lambda}, \mb{\mu}, \mb{\kappa})&= \int  \exp\paren{-E(\overline{\Theta}; \mb{\lambda}, \mb{\mu}, \mb{\kappa})}   p_{0}(d\overline{\Theta}). \label{expr: partition_fun_general}
\end{align}
The factor $\cL_{C,\Theta,\xi_{n}}\defeq \cL_{C}(\cdot; \Theta, \xi_{n})$ is defined as in \eqref{expr: error_hinge}, $\tilde{\cL}_{D,z}\defeq \tilde{\cL}_{D}(\cdot; z, \cdot)$ is defined as in \eqref{expr: entropy_constraint}. See the Appendix Sec. \ref{app: prop_1} for a detailed derivation.
\end{theorem} Moreover, we specify the error function as 
\begin{align}
\cL_{C}\paren{p, (y_{n}, \mb{x}_{n}); \Theta, \xi_{n}} &\defeq \xi_{n}- y_{n}f(\mb{x}_{n}), \label{eqn: decision_boundary}
\end{align} where $\Theta\defeq f: \cX \rightarrow \cY$ is a  decision function associated with a nonparametric classifier as defined in Sec \ref{sec 2_1}.
 \begin{figure*}[tb] 
  \centering
  \centerline{\includegraphics[width=0.85 \textwidth]{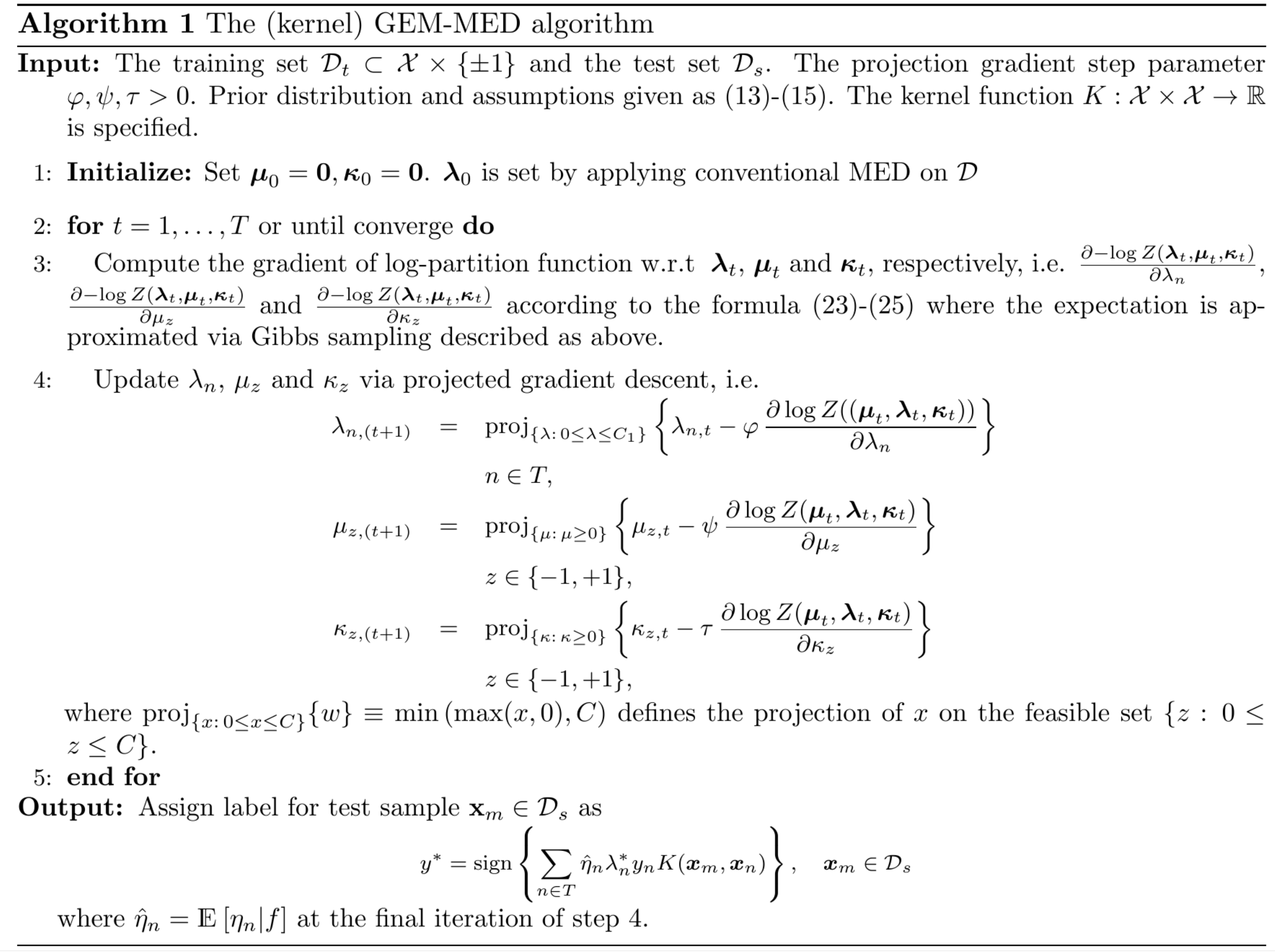}}
  \caption{\footnotesize{The proposed GEM-MED algorithm based on the projected stochastic gradient descent \cite{bertsekas1999nonlinear}. The gradient with respect to dual variables (20)-(23) can be approximated via Gibbs sampling as discussed in Sec. \ref{sec: Implement}. The constraints on the dual variable $\lambda_{n} \in [0, C_{1}]$ are imposed by a \emph{clipping} procedure $\text{proj}_{\{w:\, 0\le w \le C \}}\{w\}= \min\left(\max(w, 0),C\right)$ that is applied on each Gibbs move, similarly to the C-SVM algorithm \cite{chang2011libsvm}.  The parameters $(\psi, \varphi, \tau)$ control the stepsize of the gradient descent algorithm. } }
\label{fig: algorithm}
\end{figure*}\vspace{-5pt}
 

Since the optimization problem is convex, we can equivalently solve a dual version of the optimization problem \eqref{expr: GEM-MED}. In fact, we have the following result: 
\begin{theorem}\label{prop: dual_opt}
Assume that \eqref{expr: Gaussian_prior},  \eqref{expr: prior_factor}, \eqref{expr: Other_prior} hold, the dual optimization problem is given as 
\begin{align}
&\max_{\mb{\lambda}, \mb{\mu}, \mb{\kappa} \ge 0}  -\log Z(\mb{\lambda}, \mb{\mu}, \mb{\kappa}) \label{eqn: dual_opt}\\
&=-\log  \int  \exp\paren{-E(\overline{\Theta}; \mb{\lambda}, \mb{\mu}, \mb{\kappa})}   p_{0}(d\overline{\Theta})\nonumber\\
&= \sum_{n\in T}\paren{\lambda_{n}+ \log\paren{1- \lambda_{n}/c}}-\sum_{z\in \set{\pm 1}}\mu_{z}\hat{\gamma}_{z}+  \hat{\beta}\sum_{z\in \set{\pm 1}}\kappa_{z}\nonumber\\
&\phantom{=} -\log \int \exp\paren{ \frac{1}{2}Q(\mb{K}\odot(\mb{y}\mb{y}^{T}), \paren{\mb{\lambda}\odot \mb{\eta}}) } \nonumber\\
&\phantom{====} \times p_{0}(\mb{\eta})\exp\paren{\mb{\eta}^{T}\paren{-\mb{\mu}\otimes \mb{d} + \mb{\kappa}\otimes \mb{e}}  }d\mb{\eta}  \label{eqn: dual_opt_2}
\end{align}
where $(\mb{\lambda}, \mb{\mu}, \mb{\kappa}) $ are nonnegative dual variables as defined in \eqref{expr: opt_general},  $\mb{e}$ is the all $1$'s vector, $\odot$ is Hadamard product, $\otimes$ is the Kronecker product, respectively, and
\begin{align*}
Q(\mb{K}, \mb{x}) &= \mb{x}^{T}\mb{K}\mb{x}
\end{align*}
is the quadratic form associated with the kernel $K$.
\end{theorem}
See Appendix Sec. \ref{app: prop_2} for derivations of this result.

It is seen from \eqref{eqn: dual_opt} that the dual objective function is concave w.r.t. dual variables $(\mb{\lambda}, \mb{\mu}, \mb{\kappa})$. However, the integral in \eqref{eqn: dual_opt_2} is not closed form, so an explicit form as a quadratic optimization in SVM is not available. Nevertheless, the only coupling in \eqref{eqn: dual_opt_2} comes from the joint distribution $q(f, \mb{\eta})$. In particular, under the prior assumption   \eqref{expr: Gaussian_prior},  \eqref{expr: prior_factor}, \eqref{expr: Other_prior},  the optimal solution \eqref{expr: opt_general} satisfies
\begin{enumerate}\label{lem: 1}
\item $q(\overline{\Theta})= q(f, \mb{\eta})\prod_{n}q(\xi_{n})q(\gamma_{+1})q(\gamma_{-1})$ is factorized. 
\item $q(\mb{\eta}| f) = \prod_{n\in T}q(\eta_{n}|f)$, i.e. the $\set{\eta_{n}, n\in T}$ are conditional independent given the decision boundary function $f$. Moreover,
\begin{align}
q(\eta_{n}|f)&= Ber(q_{\eta}), \label{eqn: post_eta}\\
&\text{with } q_{\eta} = \sigma\paren{ \rho_{n}\cF_{n}(f) }\nonumber
\end{align} where
$\rho_{n} \defeq \log\frac{1- p_{0}(\eta_{n}=1)}{p_{0}(\eta_{n}=1)}$, $\cF_{n}(f) \defeq \lambda_{n}\left[ y_{n}f(\mathbf{x}_{n}) -1\right] $ $ - \mu_{y_{n}}h_{n}+ \kappa_{y_{n}}/\abs{T}, \sigma(\cdot)$ is the sigmoid function as \eqref{expr: Other_prior}.\\
\item $f| \mb{\eta} \sim \cN( f| \hat{f}_{\mb{\eta}, \mb{\lambda}}(\cdot) , \mb{K} )$, where
\begin{align}
\hat{f}_{\mb{\eta}, \mb{\lambda}}(\cdot) &= \sum_{n\in T}\lambda_{n}\eta_{n}y_{n}K(\cdot, \mb{x}_{n})\in \cH \label{eqn: post_f}
\end{align}
\end{enumerate}
See Appendix Sec. \ref{app: lemma_1} for details.

Given above results, 
we propose to use the \emph{projected stochastic gradient descent} (PSGD,\cite{bertsekas1999nonlinear, murphy2012machine}) algorithm to solve the dual optimization problem in \eqref{eqn: dual_opt_2}. The gradient vectors of the dual objective function in \eqref{eqn: dual_opt_2} w.r.t. $\mb{\lambda}$, $\mb{\mu}$, $\mb{\kappa}$, respectively, are computed as 
\begin{align}
&\partdiff{}{\lambda_{n} }  \brac{ -\log Z(\mb{\lambda}, \mb{\mu}, \mb{\kappa})} \nonumber\\
&  = 1-\E{q(f, \boldsymbol{\eta})}{ \eta_{n}y_{n}f(\mathbf{x}_{n}) }+ \frac{c}{c-\lambda_{n}}, \;  n\in T; \label{eqn: grad_lambda} \\
&\partdiff{}{\mu_{z} }  \brac{ -\log Z(\mb{\lambda}, \mb{\mu}, \mb{\kappa})} \nonumber\\
&=  \mathds{E}_{q(f, \boldsymbol{\eta})}\left\{ \sum_{n: y_{n} = z}\eta_{n}d_{n} \right\}- \hat{\gamma}_{z},\;\;  z\in \set{\pm 1}; \label{eqn: grad_mu} \\
&\partdiff{}{\kappa_{z} }  \brac{ -\log Z(\mb{\lambda}, \mb{\mu}, \mb{\kappa})} \nonumber\\
&= \hat{\beta} -  \frac{1}{\abs{T}}\E{q(f, \boldsymbol{\eta})}{\sum_{n: y_{n} = z}\eta_{n}}, \;\;  z\in \set{\pm 1}. \label{eqn: grad_kappa} 
\end{align}
Note that the expectation w.r.t. $q(f,\mb{\eta})$ are approximated by Gibbs sampling with each conditional distribution given by \eqref{eqn: post_eta}, \eqref{eqn: post_f}. For a detailed implementation of the Gibbs sampler, see the Appendix Sec. \ref{app: Gibbs}.

A complete description of algorithm is presented in \textbf{Algorithm 1}. It is remarked that in \eqref{eqn: post_eta} the probability of $\{\eta_{n}=0\}$ is proportional to the sum of margin of classification and negative local entropy value. The role of the dual variables $(\eta_{n}, \mu_{c})$ in \eqref{eqn: post_eta} and \eqref{eqn: post_f} 
is to balance the classification margin $y\,f(\cdot)$ and local entropy $h$ in determining the anomalies. 
\vspace{-15pt}

\subsection{Prediction and detection on test samples}\vspace{-10pt}
The GEM-MED classifier is similar to the standard MED classifier in \eqref{expr: test_label_MED}: 
\begin{align}
y^{*}&= \argmax_{y} \set{ \int yf(\mb{x}_{m})q(f| \hat{\mb{\eta}}, \cD_{t})df }, \nonumber\\
&=   \text{sign}\set{ \sum_{n\in T}\hat{\eta}_{n}\lambda_{n}^{*}y_{n}K(\mb{x}_{m}, \mb{x}_{n}) }   \quad \mb{x}_{m}\in \cD_{s}. \label{eqn: test_est}
\end{align}
where $\hat{\mb{\eta}}$ 
 is the conditional mean estimator of $\boldsymbol{\eta}$ given by Algorithm 1. 

The GEM-MED was optimized on the training set to detect and mitigate anomaly corrupted training samples. When there are also anomalies in the test sample, an anomaly detection method can be applied independently to screen out these samples (at a given false positive rate) before applying GEM-MED to classify them. Such a two-stage approach to handling anomalies in the test sample is obviously not optimal. An optimal joint approach to handling anomalies in the training and test samples is worthwhile future direction which will not be investigated here .

\vspace{-15pt}
\section{Experiments}
We illustrate the performance of the proposed GEM-MED algorithm on simulated data as well as on a real  data collected in a field experiment. We compare the proposed GEM-MED with the SVM implemented by \emph{LibSVM} \cite{chang2011libsvm} and the Robust-Outlier-Detection algorithm implemented with code obtained from the authors of \cite{xu2006robust}. For the simulated data experiment, a linear kernel SVM is implemented, and for the real data, a Gaussian RBF kernel SVM with  kernel $K(\mathbf{x}_{i}, \mathbf{x}_{j}) = \exp(-\gamma\| \mathbf{x}_{i} - \mathbf{x}_{j}\|_{2}^{2}) $ is implemented and the kernel parameter $\gamma>0$ is tuned via $5$-fold-cross validation. 

\begin{figure}[t] 
  \centering
\begin{minipage}[b]{0.8\linewidth}
  \centering
  \centerline{\includegraphics[scale = 0.56]{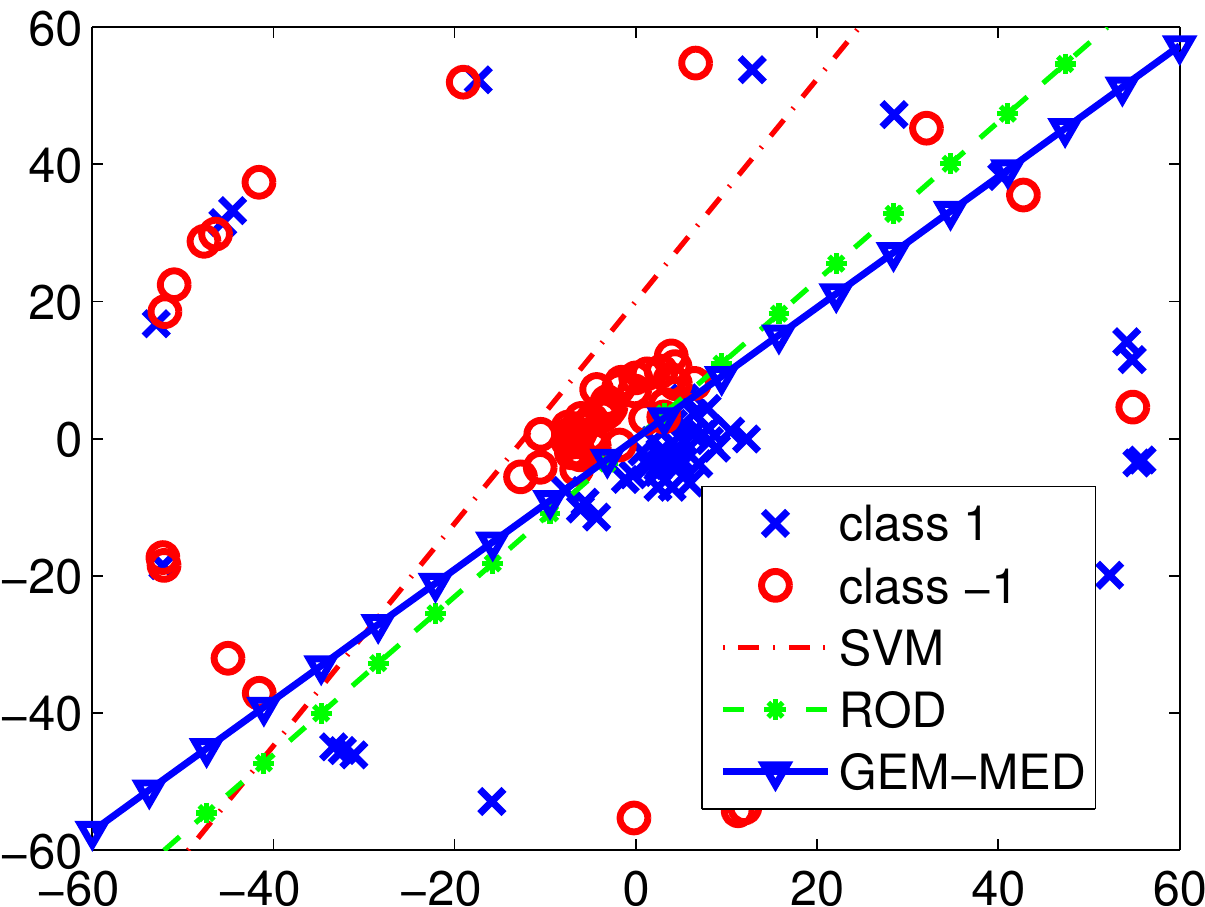}}
  \vspace{-7pt}
  \centerline{\footnotesize{(a)}}
\end{minipage}\\
\hfill
\begin{minipage}[b]{0.8\linewidth}
  \centering
   \centerline{\hspace{5pt}\includegraphics[scale = 0.5, clip=true, trim= 20mm 2mm 30mm 5mm]{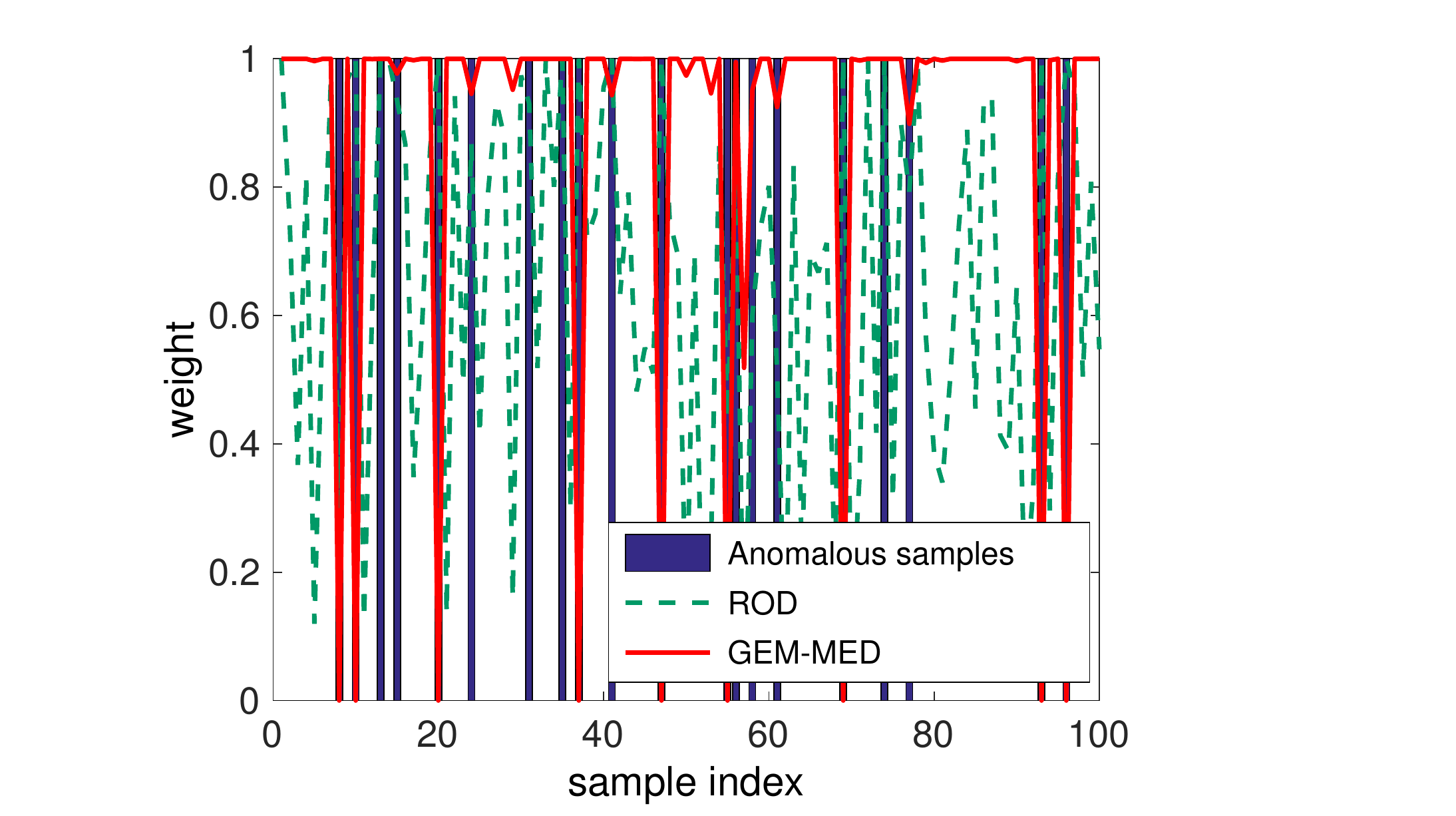}}
  \vspace{-5pt}
  \centerline{\footnotesize{(b)}}
\end{minipage}\hfill
\vspace{-10pt}
\caption{\footnotesize{(a) The classification decision boundary for SVM, ROD and GEM-MED on the
simulated data set with two bivariate Gaussian distribution $\mathcal{N}(\boldsymbol{m}_{+1}, \Sigma)$, $\mathcal{N}(\boldsymbol{m}_{-1}, \Sigma)$ in the center and a set of anomalous samples for both classes distributed in a ring. Note that SVM is biased toward the anomalies (within outer ring support) and ROD and GEM-MED are insensitive to the anomalies. (b) The Illustration of anomaly score $\hat{\eta}_{n}$ for GEM-MED and ROD. The GEM-MED is more accurate than ROD in term of anomaly detection.}}
\label{fig: experiments_sim_1}
\end{figure}

\subsection{Simulated experiment}\label{sec: exp1}
For each class $c\in\{\pm 1\}$, we generate samples from the bivariate Gaussian distribution $\mathcal{N}(\boldsymbol{m}_{+1}, \Sigma)$ and $\mathcal{N}(\boldsymbol{m}_{-1}, \Sigma)$, with mean $\boldsymbol{m}_{-1} = (3,3)$ and $\boldsymbol{m}_{+1} = -\boldsymbol{m}_{-1}$ and common covariance 
$ \Sigma = \left[\begin{array}{cc}
20 & 16 \\ 
16 & 20
\end{array} \right]. $ The sample follows the log-linear model $\log p(y, \mb{x}; \overline{\Theta})\propto 1/2\,y(\mathbf{w}^{T}\,\mathbf{x} + b)$ where $\overline{\Theta} = (\mathbf{w},b)$. A Gaussian prior was used as $p_{0}(\overline{\Theta}) = \mathcal{N}(\mathbf{w};\,\mathbf{0}, \sigma_{w}^{2}\mathbf{I})\mathcal{N}(b; 0, \sigma_{b}^{2})$. 

\begin{figure*}[t] 
\begin{minipage}{0.5\textwidth}
  \centering
  \centerline{\includegraphics[scale = 0.55, clip=true, trim= 0mm 1mm 0mm 0mm]{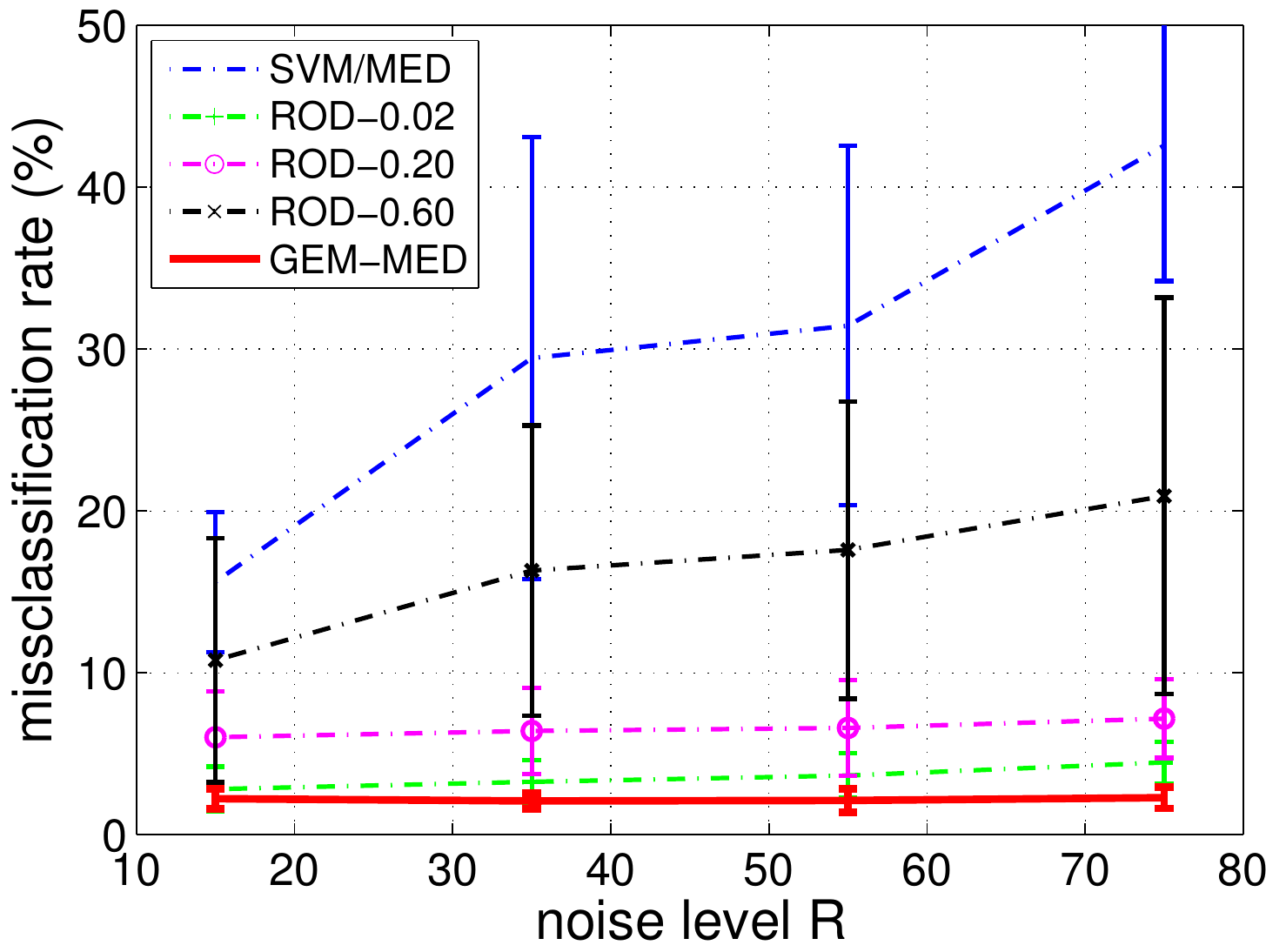}}
  \vspace{-7pt}
  \centerline{\footnotesize{(a)}}
\end{minipage}
\hfill
\begin{minipage}{0.5\textwidth}
  \centering
  \centerline{\includegraphics[scale = 0.55,  clip=true, trim= 0mm 1mm 0mm 0mm]{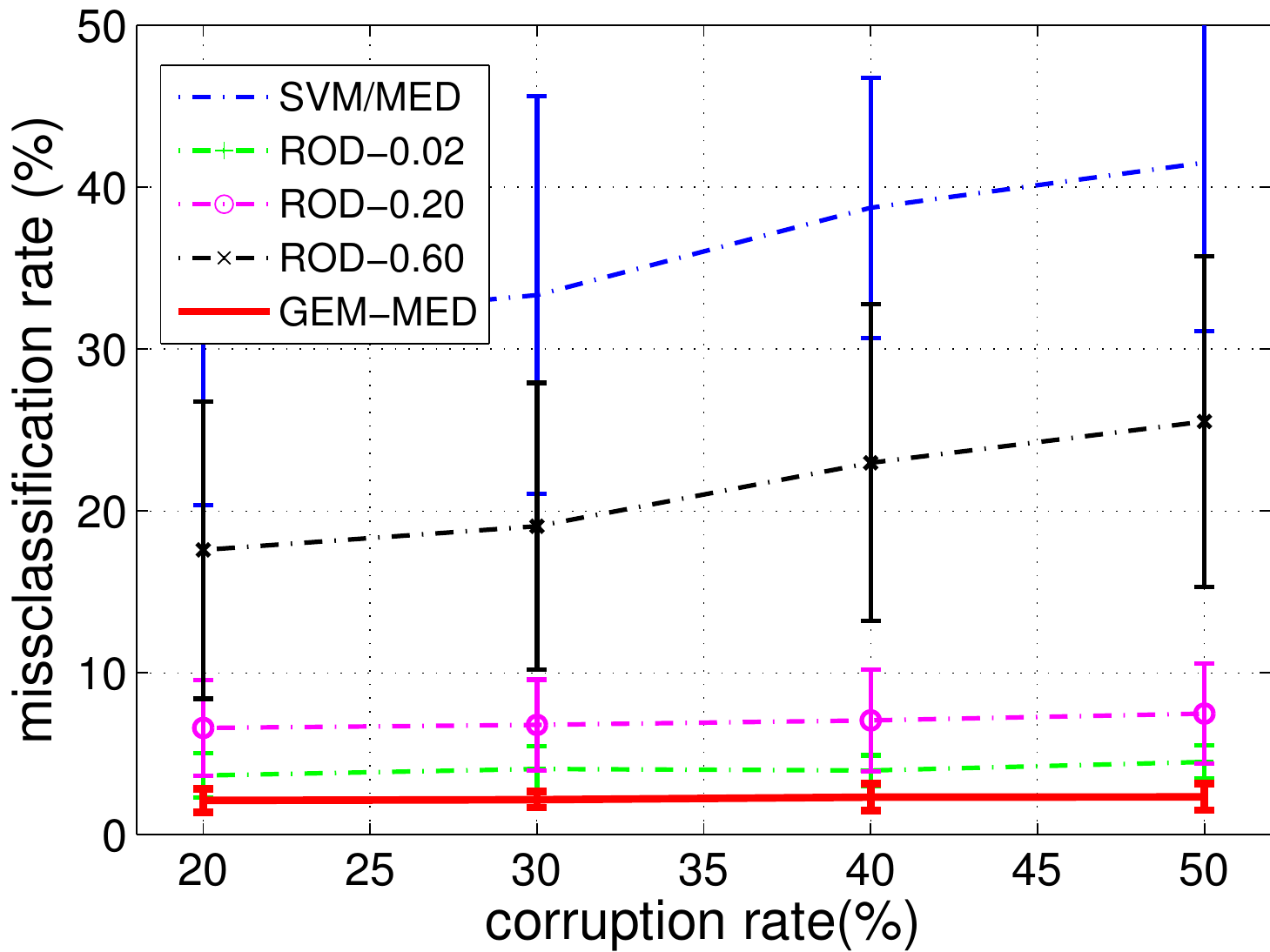}}
  \vspace{-7pt}
 \centerline{\footnotesize{(b) }}
\end{minipage}\\
\begin{minipage}{0.5\textwidth}
 \centering
  \centerline{\includegraphics[scale = 0.57]{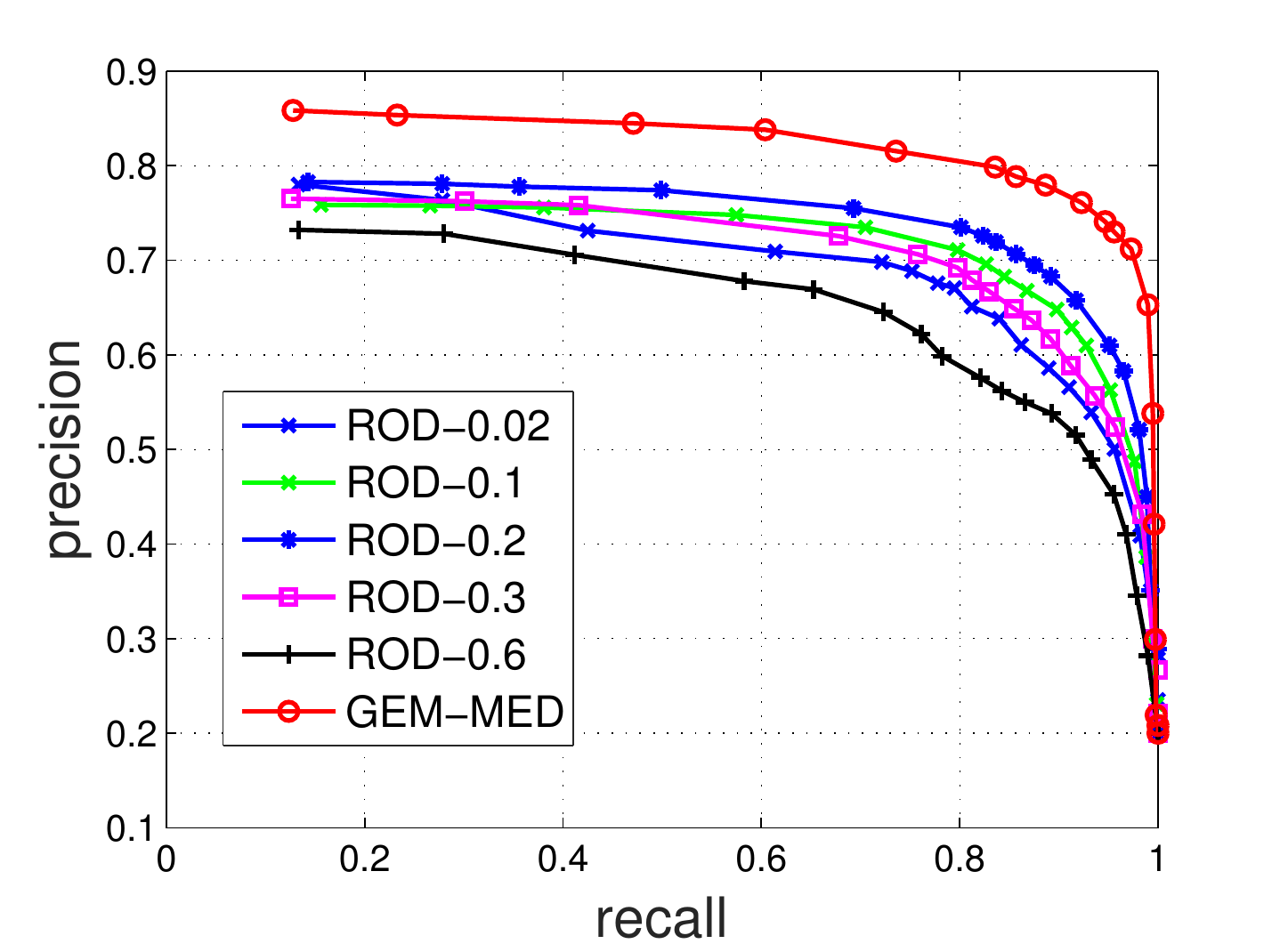}}
  \vspace{-7pt}
  \centerline{\footnotesize{(c) }}
 \end{minipage}
 \begin{minipage}{0.5\textwidth}
  \centering
  \centerline{\includegraphics[scale = 0.57, clip=true, trim= 0mm 0mm 0mm 0mm]{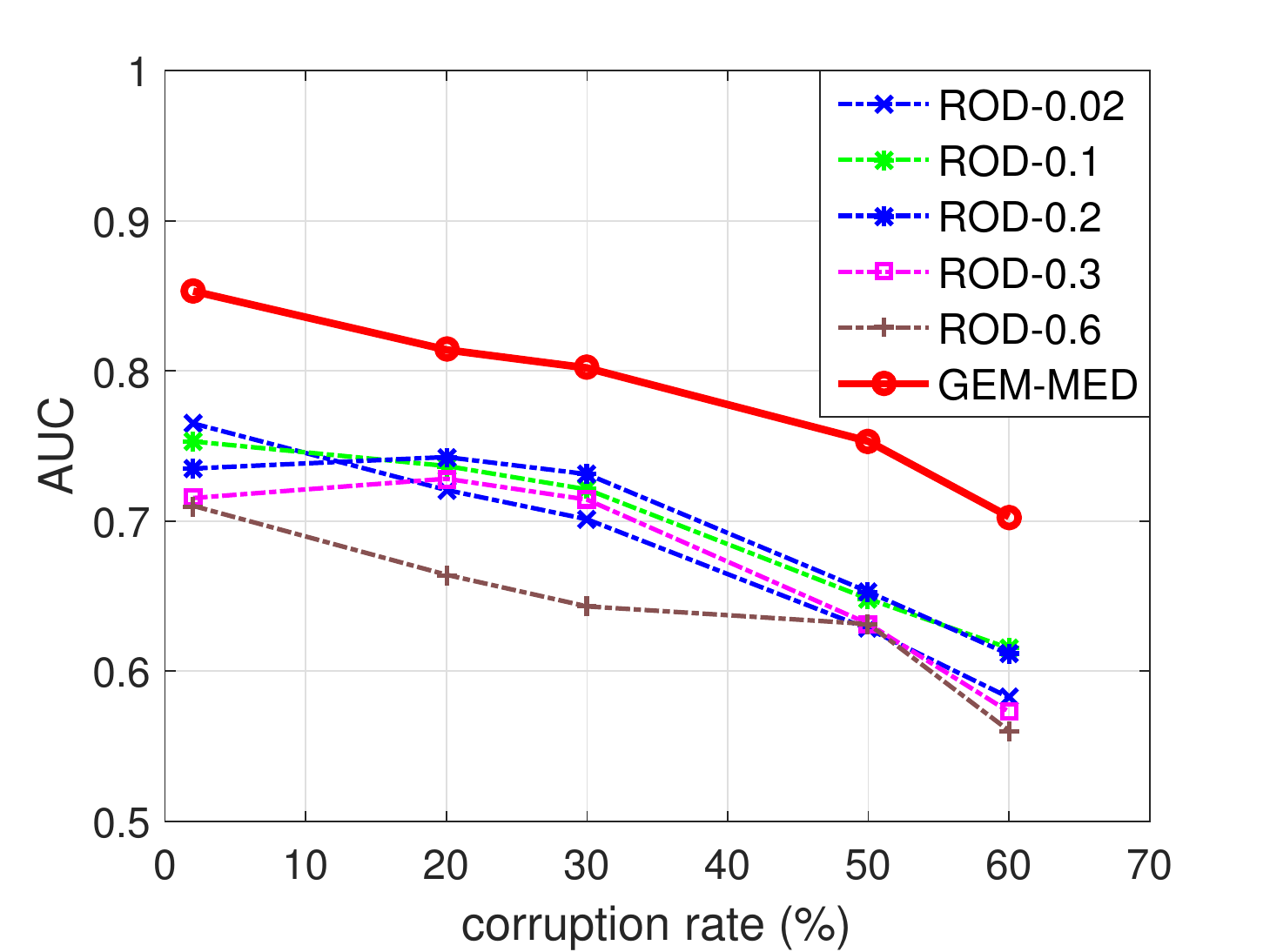}}
  \vspace{-7pt}
  \centerline{\footnotesize{(d)}}
\end{minipage}
 \vspace{-10pt}\\
\caption{\footnotesize{(a) Miss-classification error ($\%$) vs. noise level $R$ for corruption rate $r_a=0.2$. (b) Miss-classification error ($\%$) vs. corruption rate $\E{}{\eta}$ for ring-structureed anomaly distribution having ring  $R=55$. (c) Recall-precision curve for GEM-MED and RODs on simulated data for corruption rate $=0.2$. (d) The AUC vs. corruption rate $r_a$ for GEM-MED and ROD with a range of outlier parameters $\rho$. From (a) and (b),  GEM-MED  outperforms both SVM/MED and ROD for various $\rho$ in classification accuracy. From (c), under the same corruption rate, we see that GEM-MED outperforms ROD in terms of the precision-recall behavior. This due to the superiority of GEM constraints in enforcing anomaly penalties into the classifier. From (d),  The GEM-MED outperforms RODs in terms of AUC for the range of investigated corruption rates. }}
\label{fig:pred-acc-prec-recall}\vspace{-5pt}
\end{figure*}

We followed the same models as in \cite{xu2006robust}. In particular, the anomalies in the training set were drawn uniformly from a ring with an inner radius of $R$ and outer radius $R+1$, where $R$ was assigned as one of the values $[15, 35, 55, 75]$.  Define $R$ to be the \emph{noise level} of the data set, since the larger $R$ the higher the discrepancy between the nominal distribution and the anomalous distribution. 
The samples then were labeled as $\{0,1\}$ with equal probability.  The size of the training set was $100$ for each class, and the ratio of anomaly samples was $r_{a}$. The test set contained $2000$ uncorrupted samples from each class. See Fig. \ref{fig: experiments_sim_1} (a) for a realization of the data set and the classifiers.
 


\begin{figure*}[htb] 
\begin{minipage}{0.5\textwidth}
  \centering
  \centerline{\includegraphics[scale = 0.57, clip=true, trim= 0mm 0mm 0mm 0mm]{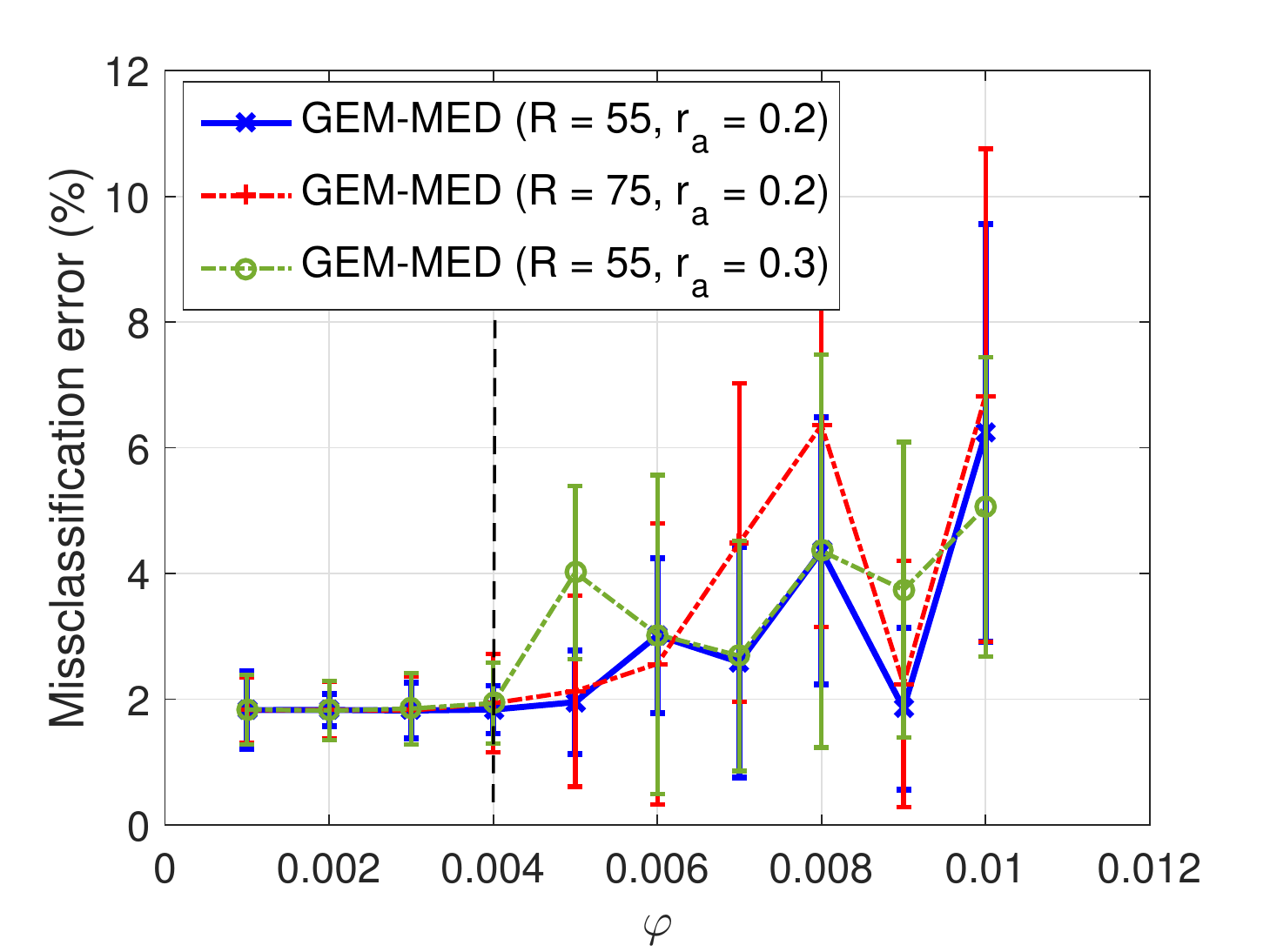}}
  \vspace{-7pt}
  \centerline{\footnotesize{(a)}}
\end{minipage}
\begin{minipage}{0.5\textwidth}
  \centering
  \centerline{\includegraphics[scale = 0.57, clip=true, trim= 0mm 0mm 0mm 0mm]{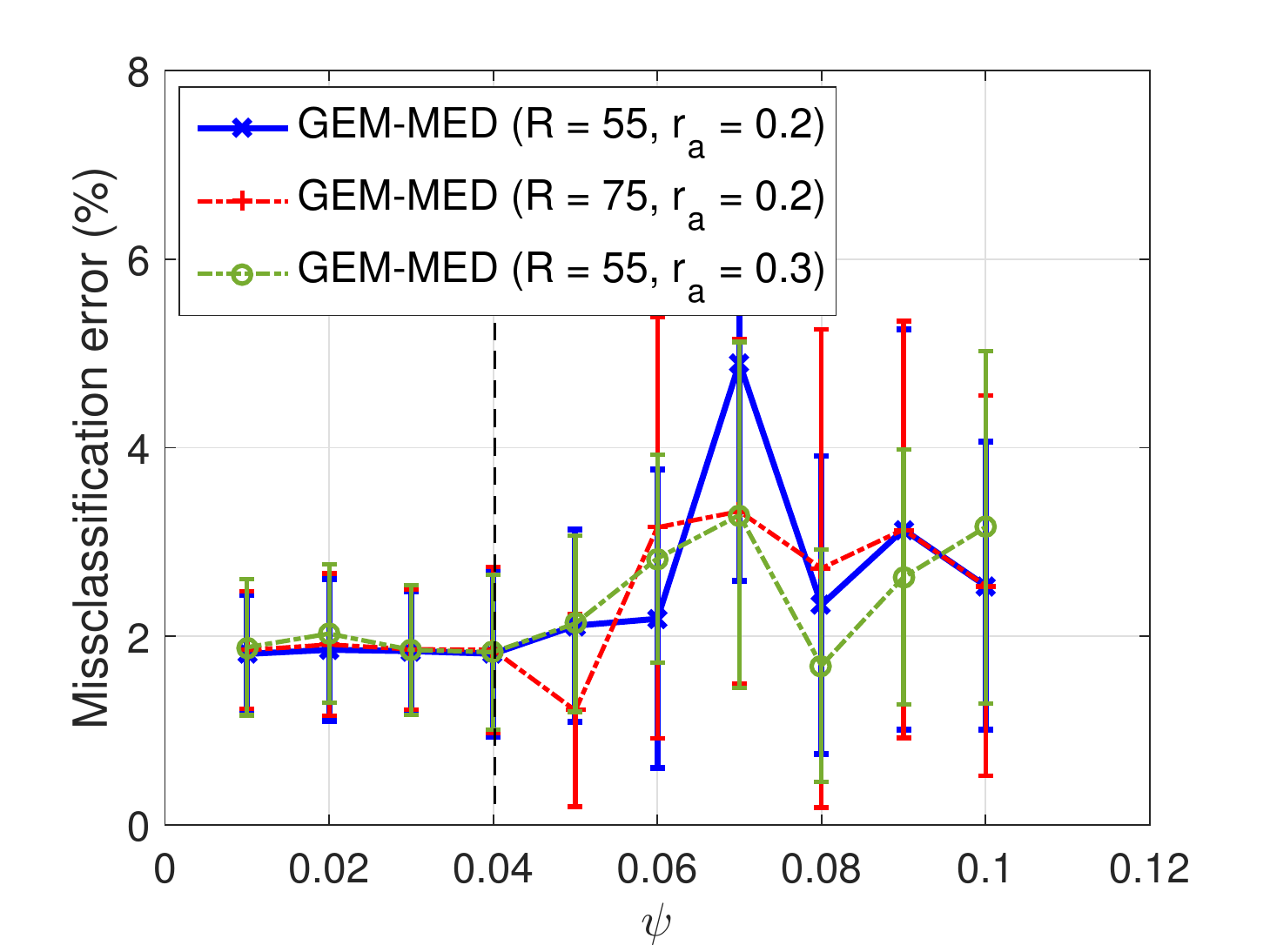}}
  \vspace{-7pt}
  \centerline{\footnotesize{(b)}}
\end{minipage}\\
{\centering
\begin{minipage}{1\textwidth}
  \centering
  \centerline{\includegraphics[scale = 0.57, clip=true, trim= 0mm 0mm 0mm 0mm]{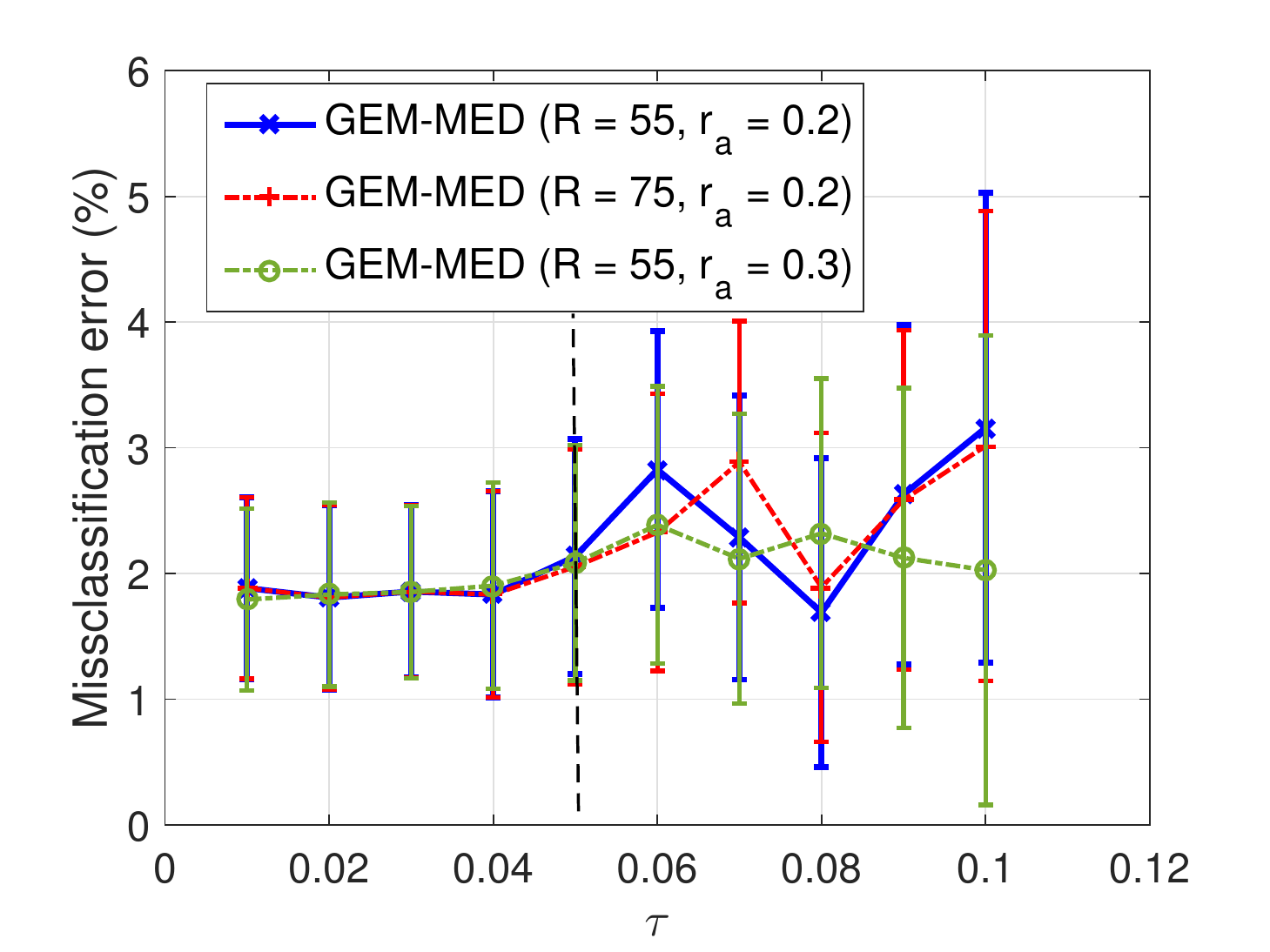}}
  \vspace{-7pt}
  \centerline{\footnotesize{(c)}}
\end{minipage}\hfill}
\caption{\footnotesize{The classification error of GEM-MED vs. (a) learning rates $\varphi$, when $(\psi = 0.01, \tau = 0.02)$; (b) vs. $\psi$ when $(\varphi = 0.001, \tau = 0.02)$ and (c) vs. $\tau$ when $(\varphi = 0.001, \psi = 0.01)$. 
 The vertical dotted line in each plot separates the breakdown region (to the right) and the stable region of misclassification performance. These threshold values do not vary significantly as the noise level $R$ and corruption rate $r_a$ vary over the ranges investigated.  }}
\label{fig: auc_rate}\vspace{-5pt}
\end{figure*}

We first compare the classification accuracy of SVM, Robust-Outlier-Detection (ROD) with outlier parameter $\rho$ and GEM-MED, under noise level $R$ and a range of corruption rates $r_{a} \in \set{0.2, 0.3, 0.4, 0.5}$.  We used the BP-kNNG implementation of GEM, where the k-nearest neighbor parameter $k=5$. In the update of the GEM-MED dual variables $(\boldsymbol{\lambda, \mu, \kappa})$, the learning rate $(\varphi, \psi, \tau)$ is chosen based on a comparison of classification performance of the GEM-MED under a range of noise levels $R$ and corruption rates $r_a$, as shown in Fig. \ref{fig: auc_rate} (a)-(c). Note that when $\varphi \in [1,4] \times 10^{-3}, \psi \in [1,4] \times 10^{-2}, \tau \in [1,5] \times 10^{-2}$,  the performance of the GEM-MED is stable in terms of the averaged missclassification error and the variance. We fix $(\varphi, \psi, \tau)$ in the stable range in the following experiments. 
For the ROD, we investigated a range of algorithm parameters, in particular outlier parameter $\rho\in\{0.02, 0.2,0.6\}$ for comparison, and we observed that the value  $\rho =0.02$ gives the best classification performance regardless of  the setting of $R \in \set{15, 35, 55, 75} $ or $r_a \in \set{0.2, 0.3, 0.4, 0.5}$.  Recall that the ROD parameter $\rho$ is a fixed threshold that determines the proportion of anomalies, i.e., the proportion of nonzero $\eta_{n}$ \cite{xu2006robust}. Compared to the ROD, the GEM-MED as a Bayesian method requires no tuning parameter to control the proportion of anomalies. In the experiments below, we compare the ROD for a range of outlier parameters $\rho$ with GEM-MED for a single choice of $(\varphi, \psi, \tau)$, which were tuned via 5-fold-cross-validation of misclassification rate over $50$ trial runs.

Fig. \ref{fig:pred-acc-prec-recall}(a) shows the miss-classification error ($\%$) versus various noise level R (with $r_{a}=0.2$), and Fig. \ref{fig:pred-acc-prec-recall}(b) shows the miss-classification error under different corruption rate settings (with $R =55$). In both experiments,  GEM-MED outperforms ROD and SVM in terms of classification accuracy. Note that when  the noise level or the corruption rate increases, the training data become less representative of the test data and the difference between their distributions increases. This causes a significant performance deterioration for the SVM/MED method, which is demonstrated in Fig. \ref{fig:pred-acc-prec-recall}(a) and Fig. \ref{fig:pred-acc-prec-recall}(b).  Fig. \ref{fig: experiments_sim_1} (b) also shows the bias of the SVM classifier towards the anomalies that lie in the ring.   Comparing to GEM-MED and ROD in Fig \ref{fig:pred-acc-prec-recall}(a) and Fig. \ref{fig:pred-acc-prec-recall}(b), 
the former method is less sensitive to the anomalies. Moreover, since the GEM-MED model takes into account the marginal distribution for the training sample, it is more adaptive to anomalies in the training set, as compared to ROD, which does not use any prior knowledge about the nominal distribution but only relies on the predefined outlier parameter $\rho$ to limit the training loss.   \hspace{-15pt}


In Fig. \ref{fig:pred-acc-prec-recall}(c) we compare the performance of GEM-MED and ROD in terms of precision vs recall for the same corruption rate as in Fig. \ref{fig:pred-acc-prec-recall}(a) and \ref{fig:pred-acc-prec-recall}(b).  In ROD  and GEM-MED, the estimated weights $\eta_{n} \in [0,1]$ for each sample can be used to infer the likelihood of anomalies. In particular,  in GEM-MED the corresponding latent variable estimate $\hat{\eta}_{n}$ is obtained at the final iteration of the Gibbs sampling procedure, as described in Appendix Sec. \ref{app: Gibbs}.  Following the anomaly ranking  procedure in \cite{xu2006robust}, these anomaly scores are placed in ascending order. We compute the precision and recall using this ordering by averaging over $50$ runs. Precision and recall are measures that are commonly used in data mining \cite{japkowicz2011evaluating}: 
\begin{flalign*}
 \text{Precision} &= \frac{|\{n: \eta_{n} \le \rho_{c} \} \cap \{ n: (\mathbf{x}_{n}, y_{n} ) \text{ are anomalous}\}|}{|\{n: \eta_{n} \le \rho_{c} \}|} \\
\text{Recall} 
\quad&= \frac{|\{n: \eta_{n} \le \rho_{c} \} \cap \{ n: (\mathbf{x}_{n}, y_{n} ) \text{ are anomalous}\}|}{|\{ n: (\mathbf{x}_{n}, y_{n} ) \text{ are anomalous}\}|},
\end{flalign*}  where the threshold $\rho_{c}$ is a cut-off threshold that is swept over the interval $[0,1]$ to trace out the precision-recall curves in Fig. \ref{fig:pred-acc-prec-recall}(c). It is evident from the figure that the proposed GEM-MED  outlier resistant classifier  has better precision-recall performance than ROD. Other corruption rates $r_a$ lead to similar results. In Fig. \ref{fig:pred-acc-prec-recall}(d),  we compare the performance of GEM-MED, RODs under different corruption rates
in terms of the Area Under the Curve (AUC), a commonly used measure in data mining  \cite{japkowicz2011evaluating}. Similar to Fig.\ref{fig:pred-acc-prec-recall}(c), the GEM-MED outperforms RODs in terms of AUC for the range of investigated corruption rates. 


\vspace{-5pt}
\subsection{Footstep classification experiment}
\begin{figure}[t] 
  \centering
   \includegraphics[width=6.5cm]{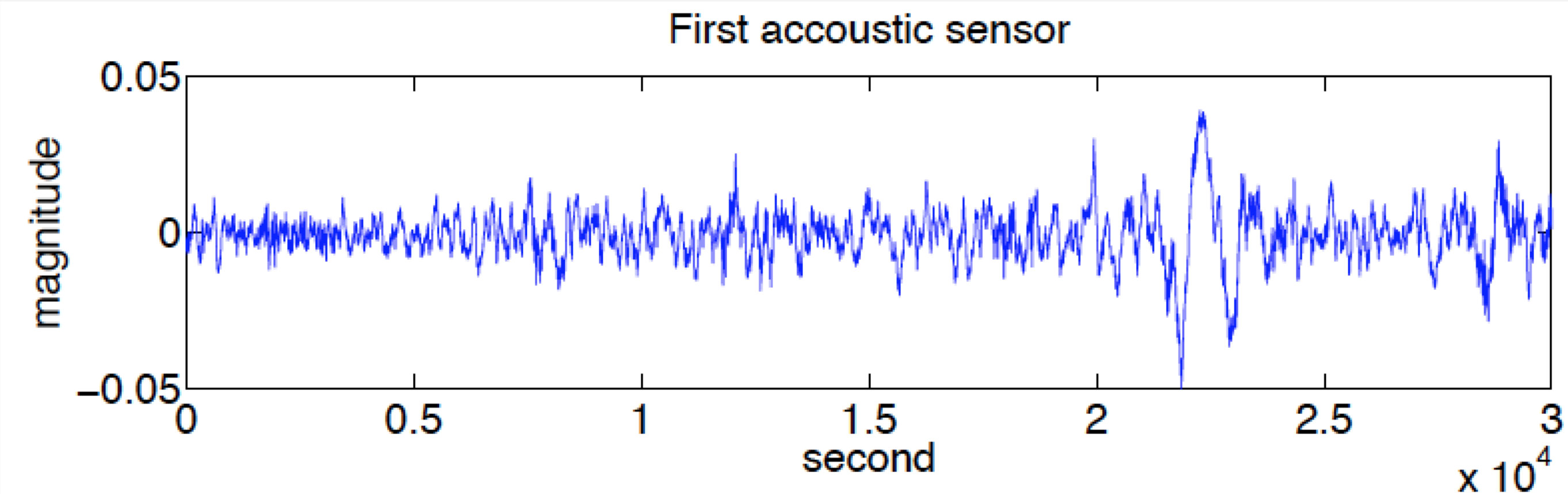}\\
   \includegraphics[width=6.5cm]{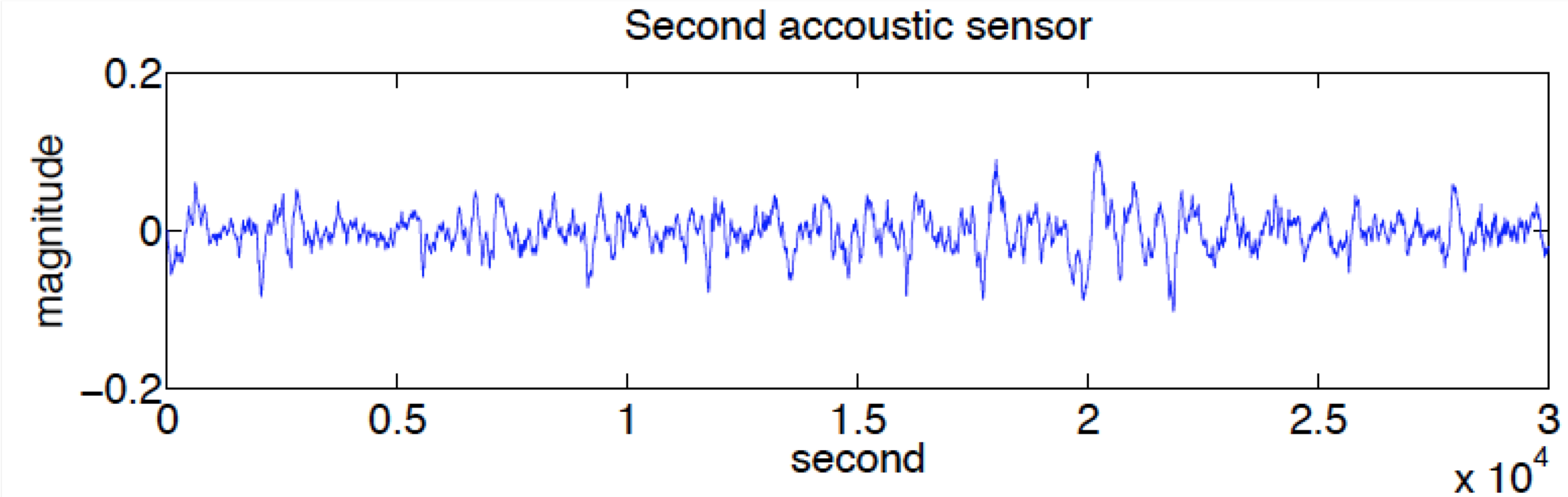}\\
   \includegraphics[width=6.5cm]{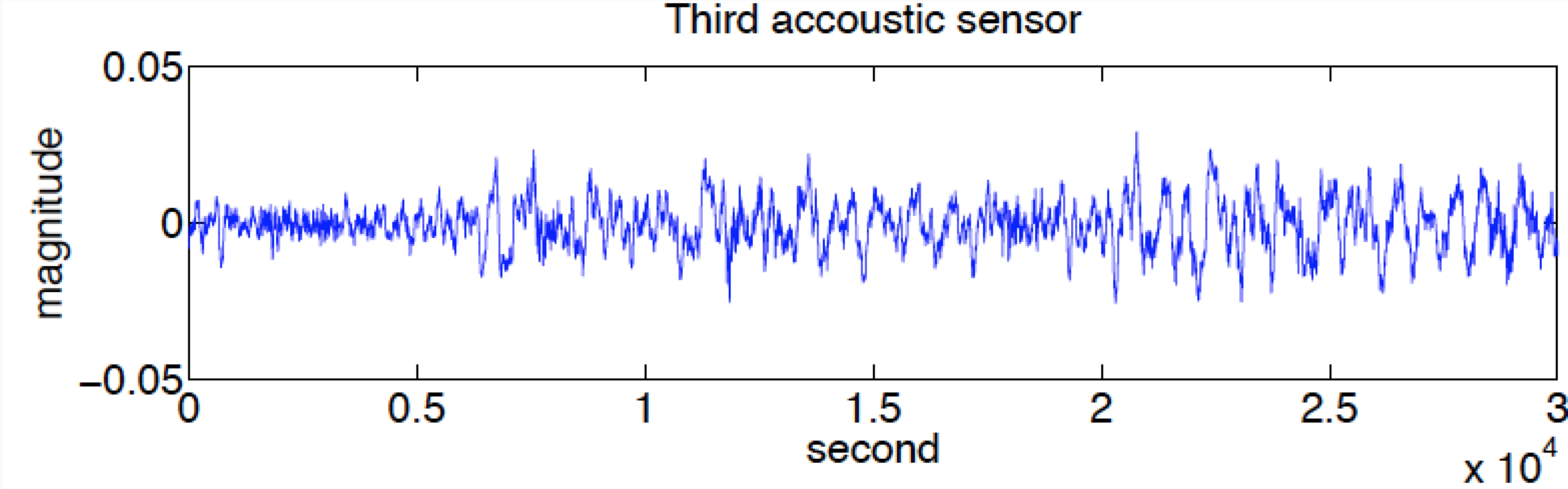}\\
   \includegraphics[width=6.5cm]{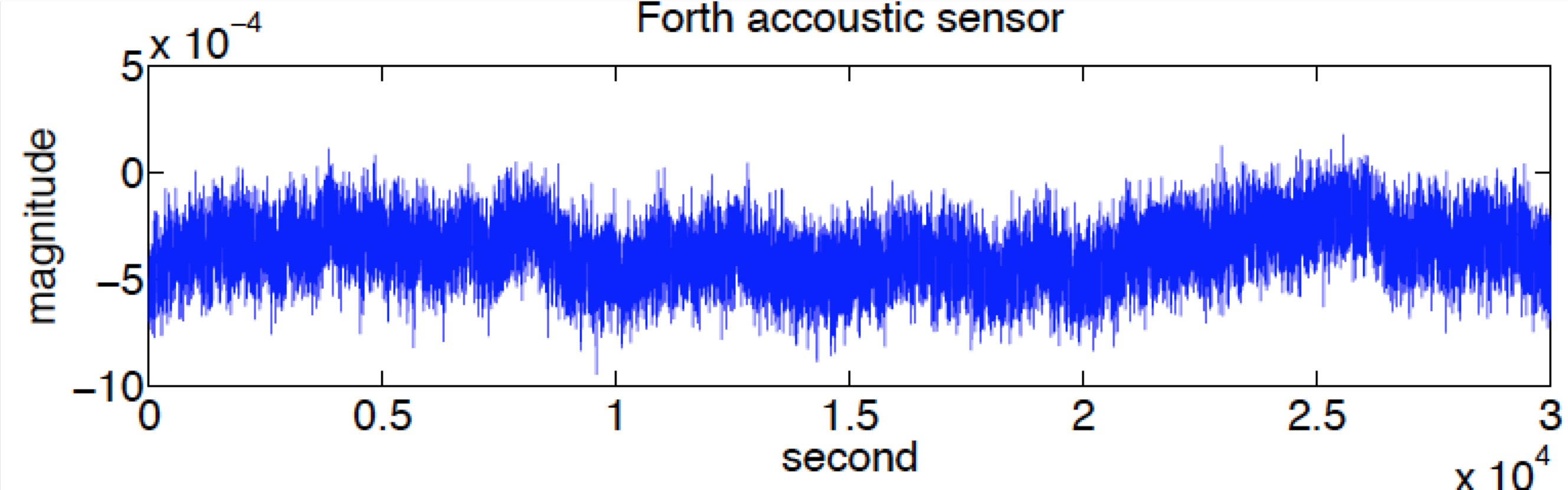}\vspace{-5pt}
\caption{\footnotesize{ A snapshot of human-alone footstep collected by four acoustic sensors.   }}
\label{fig: footstep_wave}
\end{figure}

The proposed GEM-MED method was evaluated on experiments on a real data set collected by the U.S. Army Research Laboratory  \cite{huang2011multi,damarla2012seismic,nguyen2011robust}. This data set contains footstep signals recorded by a multisensor system, which includes four acoustic sensors and three seismic sensors. All the sensors are well-synchronized and operate in a natural environment, where the acoustic signal recordings are corrupted by environmental noise and intermittent sensor failures. The task is to discriminate between human-alone footsteps and human-leading-animal footsteps. We use the signals collected via four acoustic sensors (labeled sensor 1,2,3,4) to perform the classification. 
See Fig. \ref{fig: footstep_wave}. 
Note that the fourth acoustic sensor suffers from sensor failure, as evidenced by its very noisy signal record  (bottom panel of Fig. \ref{fig: footstep_wave}). 
The data set involves $84$ human-alone subjects and $66$ human-leading-animal subjects. Each subject contains $24$ $75\%$-overlapping sample segments to capture temporal localized signal information. We randomly selected $25$ subjects with $600$ segments from each class as the training set. The test set contains the rest of the subjects. In particular, it contains $1416$ segments from human-alone subjects and $984$ segments from human-leading-animal subjects. A more detailed description of the dataset is given in \cite{huang2011multi, damarla2012seismic}.

\begin{figure}[htb] 
  \centering
  \begin{minipage}[b]{0.6\linewidth}
  \centering
  \centerline{\includegraphics[width=6.8cm, clip=true, trim= 10mm 1mm 10mm 5mm]{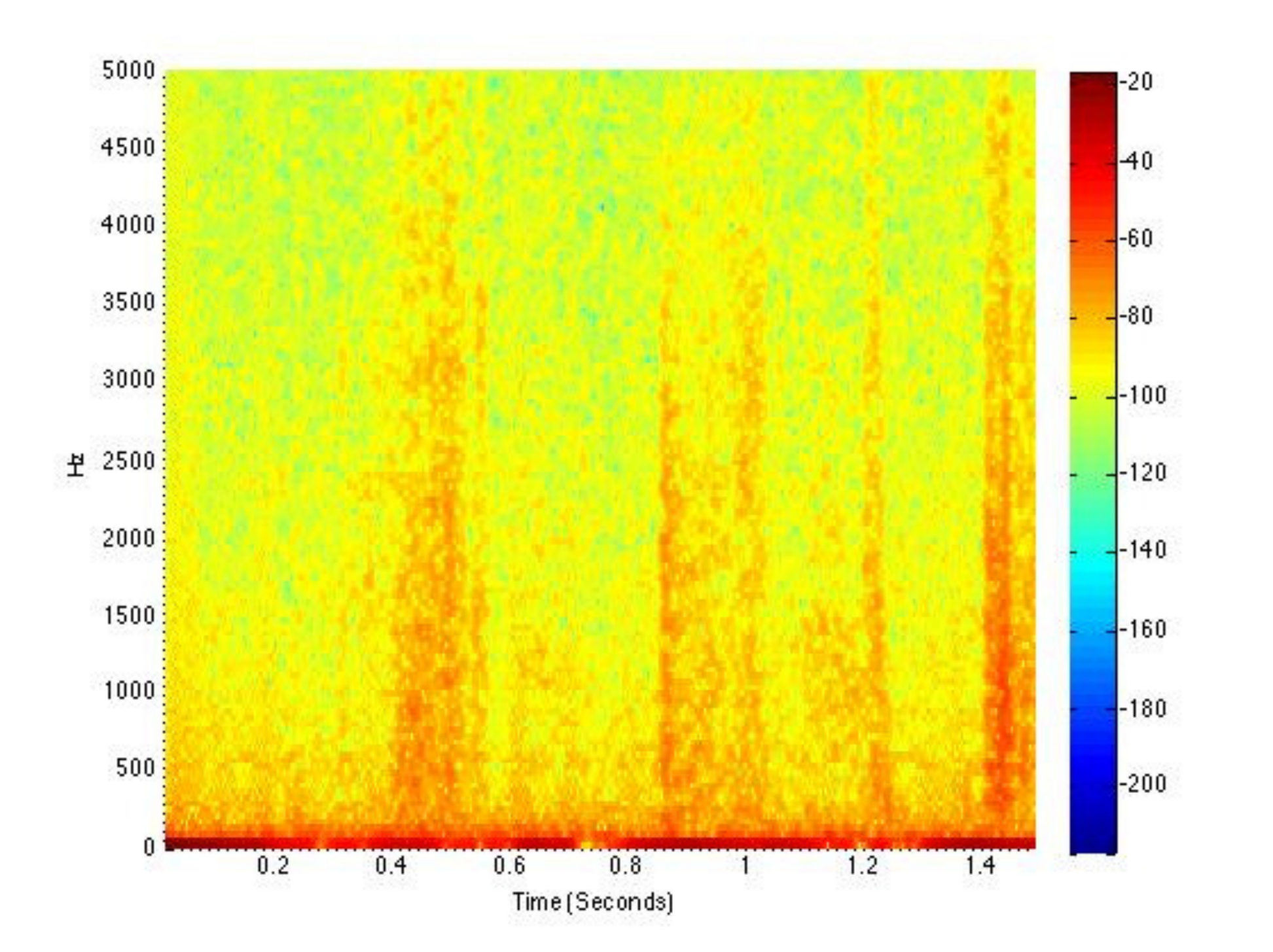}}
  \vspace{-8pt}
  \centerline{\scriptsize{(a)}}
\end{minipage}\\
\begin{minipage}[b]{0.6\linewidth}
  \centering
  \centerline{\includegraphics[width=6.8cm, clip=true, trim= 10mm 1mm 10mm 5mm]{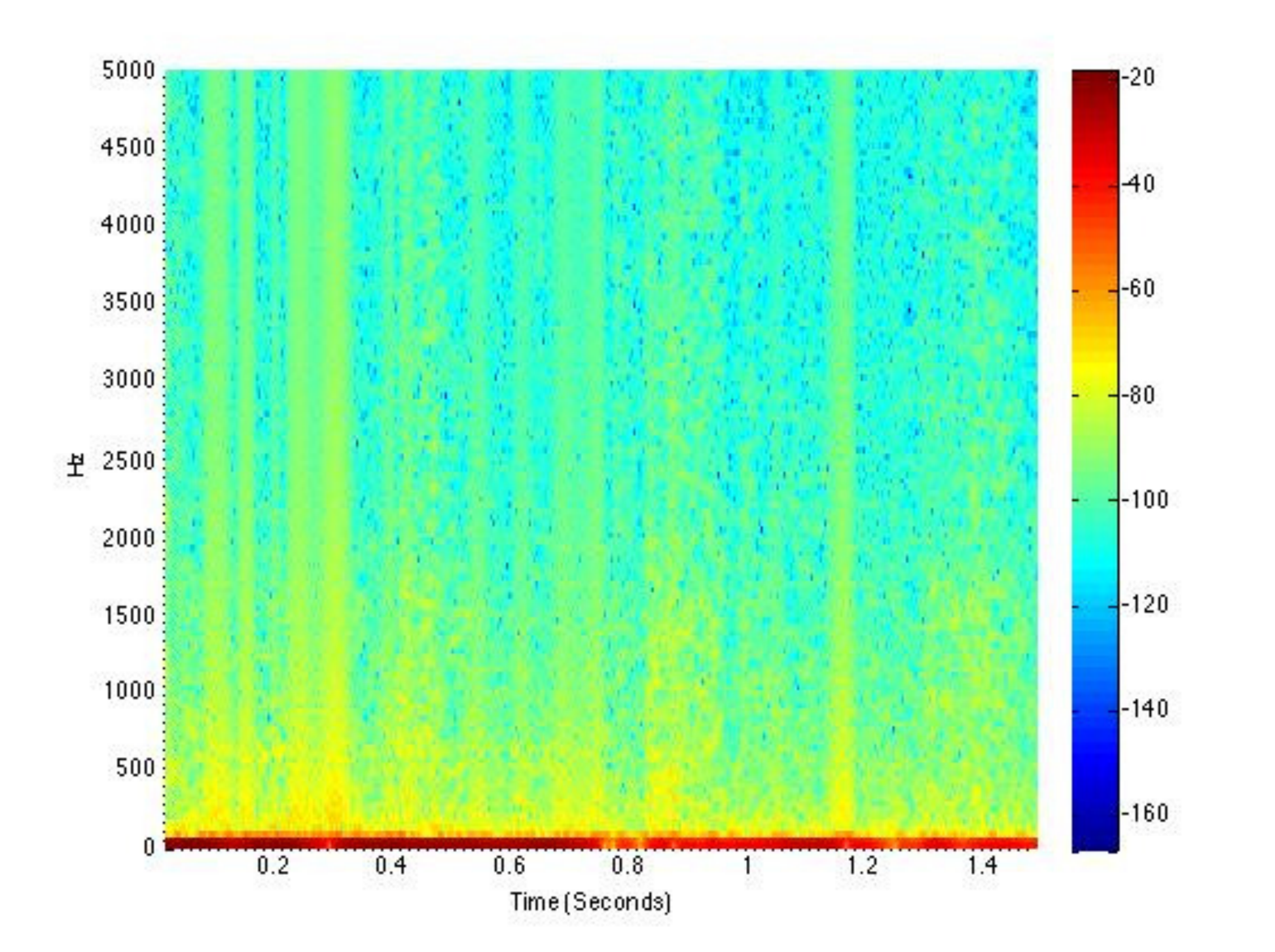}}
  \vspace{-8pt}
  \centerline{\scriptsize{(b)}}
\end{minipage} \\
\begin{minipage}[b]{0.6\linewidth}
  \centering
  \centerline{\includegraphics[width=6.8cm, clip=true, trim=0mm 1mm 10mm 5mm]{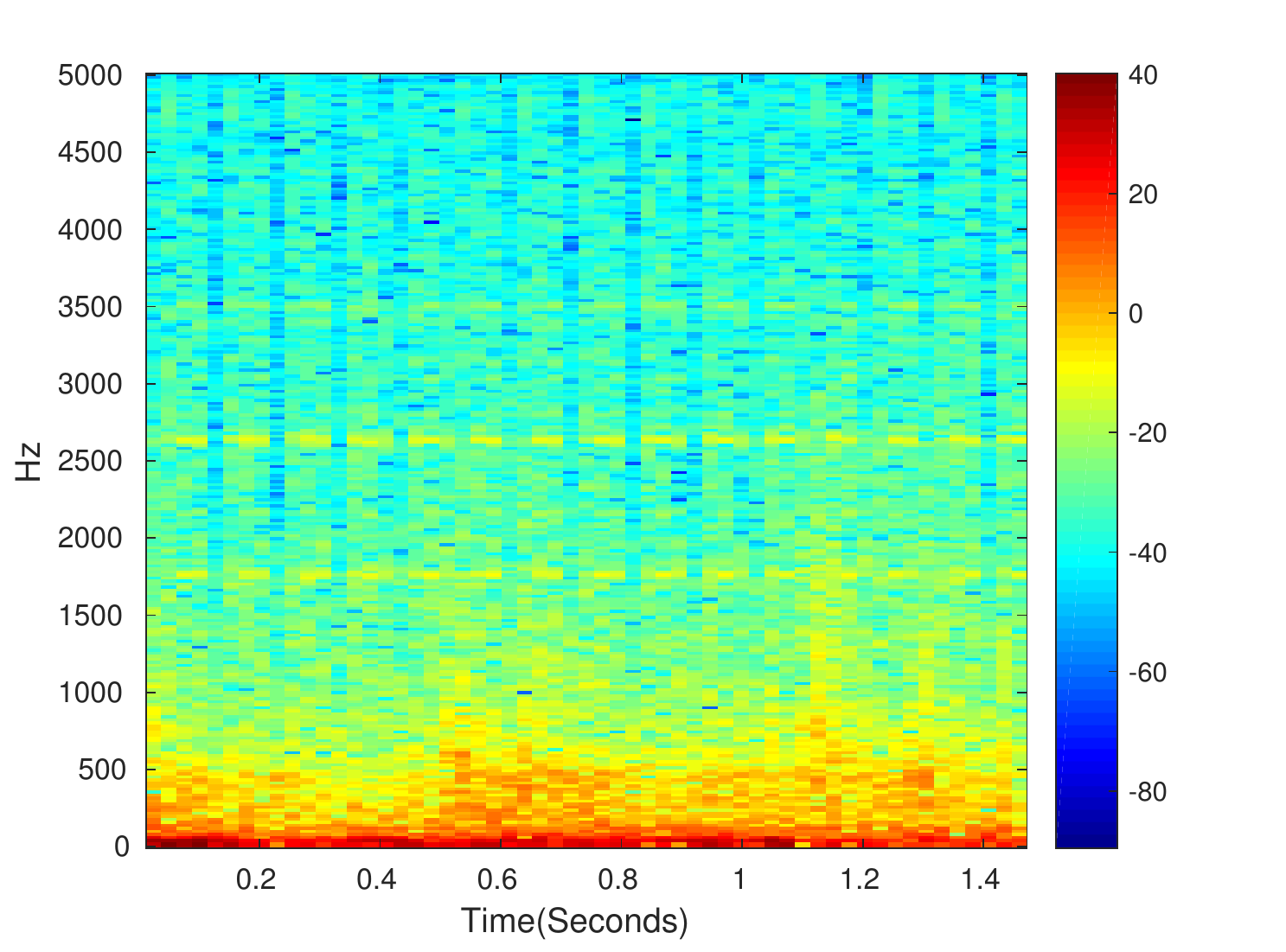}}
  \vspace{-8pt}
  \centerline{\scriptsize{(c)}}
\end{minipage}\vspace{-10pt}
\caption{\footnotesize{ The power spectrogram (dB) vs. time (sec.) and frequency (Hz.) for a human-alone footstep (a) and a human-leading-animal footstep (b).  Observe that the period of periodic footstep is a discriminative feature that separates these two signals. (c) shows a corrupted human-alone footstep due to sensor malfunctioning.  }}
\label{fig: powspec}
\end{figure}

 In a preprocessing step, for each  segment, the time interval with strongest signal response is identified and signals within a fixed size of window ($1.5 $ second) are extracted from the background.  
Fig. \ref{fig: powspec} shows the spectrogram (dB) of human-alone footsteps and human-leading-animal footsteps using the short-time Fourier transform \cite{sejdic2009time}, as a function of time (second) and frequency (Hz). The majority of the energy is concentrated in the low frequency band and the footstep periods differ between these two classes of signals. For features, we extract a mel-frequency cepstral coefficient (MFCC, \cite{mermelstein1976distance}) vector  
using a $50$ msec. window. Only the first $13$ MFCC coefficients were retained, which were experimentally determined to  capture an average $90\%$ of the power in the associated cepstra. There are in total $150$ windows for each segment, resulting in a matrix of MFCC coefficients of size $13 \times 150$. We  reshaped the matrix of MFCC features to obtain a $1950$ dimensional feature vector  for each segment. We then apply PCA to reduce the dimensionality from $1950$ to $50$, while preserving $85\%$ of the total power. The above procedures for preprocessing follows exactly from \cite{nguyen2011robust}.

\begin{table*}[htb]
\renewcommand{\arraystretch}{2.0}
\centering
\caption{\footnotesize Classification accuracy \textbf{on nominal (clean) test set} for footstep experiment  with different sensor combinations, with the best performance shown in \textbf{bold}.} 
\label{tab: class_accuracy_norm}
\begin{tabular*}{0.885\textwidth}{|m{45pt}<{\centering}|m{50pt}<{\centering}|m{55pt}<{\centering}|m{55pt}<{\centering}|m{55pt}<{\centering}|m{55pt}<{\centering}|m{55pt}<{\centering}|}
\hline 
\multicolumn{7}{|c|}{Classification Accuracy ($\%$) mean $\pm$ standard error} \\[3pt] \hline
sensor no.&  kernel SVM  & kernel MED  & ROD-$0.02$ & ROD-$0.2$ &  GEM $+$ SVM    & GEM-MED \\
\hline 
$1$ & $71.2 \pm 8.2$ &$71.1\pm 5.3$ & $73.7 \pm 3.7$ & $76.0 \pm 2.5$
 &  $72.5 \pm 4.2$  & $\mathbf{78.4 \pm 3.3}$ \\
\hline 
$2$ & $60.8 \pm 12.5 $ &$62.3 \pm 10.2$  & $71.5 \pm 7.3$ & $76.5 \pm 5.3$ 
&  $ 70.3 \pm 2.5 $  & $\mathbf{82.1 \pm 3.1}$ \\
\hline 
$3$ & $60.5 \pm 14.2 $ &$60.0 \pm 13.1$ & $63.2 \pm 5.4$ & $\mathbf{67.6 \pm 4.2}$
& $56.5 \pm 3.5$ & $66.8 \pm 4.5$ \\
\hline 
$4$ & $59.6 \pm 10.1 $  &$58.4 \pm 8.2$ & $71.8 \pm 7.2$ & $73.2 \pm 4.2$ 
&  $ 76.5 \pm 2.7 $  &$\mathbf{80.1 \pm 3.1}$ \\
\hline 
1,2,3,4 &  $75.9 \pm 7.5$ & $78.6 \pm 5.1$ & $79.2 \pm 3.7$ & $79.8 \pm 2.5$
&  $75.2 \pm 3.3$   & $\mathbf{84.0 \pm 2.3}$ \\
\hline 
\end{tabular*} \vspace{3pt}
\hfill 
\end{table*}

\begin{table*}[htb]
\renewcommand{\arraystretch}{2.0}
\centering
\caption{\footnotesize Classification accuracy \textbf{on the entire (corrupted) test set} for footstep experiment  with different sensor combinations, with the best performance shown in \textbf{bold}.} 
\label{tab: class_accuracy}
\begin{tabular*}{0.885\textwidth}{|m{45pt}<{\centering}|m{50pt}<{\centering}|m{55pt}<{\centering}|m{55pt}<{\centering}|m{55pt}<{\centering}|m{55pt}<{\centering}|m{55pt}<{\centering}|}
\hline 
\multicolumn{7}{|c|}{Classification Accuracy ($\%$) mean $\pm$ standard error} \\[3pt] \hline
sensor no.&  kernel SVM  & kernel MED  & ROD-$0.02$ & ROD-$0.2$ &  GEM $+$ SVM    & GEM-MED \\
\hline 
$1$ & $65.2 \pm 10.6$ &$65.8\pm 10.2$ & $68.5 \pm 8.3$ & $70.0 \pm 6.8$
 &  $70.2 \pm 5.5$  & $\mathbf{72.5 \pm 4.8}$ \\
\hline 
$2$ & $54.9 \pm 11.8 $ &$55.2 \pm 11.0$  & $63.2 \pm 9.8$ & $68.1 \pm 7.5$ 
&  $ 68.5 \pm 7.8 $  & $\mathbf{76.3 \pm 3.9}$ \\
\hline 
$3$ & $50.7 \pm 10.0 $ &$52.0 \pm 10.5$ & $56.8 \pm 8.5$ & $56.9 \pm 7.3$
& $56.5 \pm 3.5$ & $\mathbf{60.1 \pm 5.3}$ \\
\hline 
$4$ & $57.0 \pm 12.3 $  &$57.5 \pm 12.1$ & $69.6 \pm 9.2$ & $69.8 \pm 5.1$ 
&  $ 70.2 \pm 4.2 $  &$\mathbf{75.0 \pm 4.0}$ \\
\hline 
1,2,3,4 &  $70.8 \pm 8.8$ & $71.0 \pm 8.5$ & $73.6 \pm 7.2$ & $74.8 \pm 6.9$
&  $75.1 \pm 3.3$   & $\mathbf{76.8 \pm 2.5}$ \\
\hline 
\end{tabular*} \vspace{3pt}
\hfill 
\end{table*}


We compare the performance of kernel SVM,  kernel MED,  ROD for outlier parameter $\rho\in [0.01,1]$, and GEM-MED by training on the  four  sensors individually as well as in combination. For the combined sensors we used  an augmented feature  vector of dimension $200$ via feature concatenation.  
We used a Gaussian RBF kernel function for the matrix $\mathbf K$ in the Gaussian process prior for the SVM decision function $f$.  For the optimization of GEM-MED we used a separable prior and exponentially distributed hyperparameters, as indicated by  \eqref{expr: prior_factor} and \eqref{expr: Other_prior}.  
Finally, the BP-kNNG implementation of GEM was applied on the training samples in the MFCC feature space  
with $k=10$  nearest neighbors. The threshold $\vartheta$ is set using the Leave-One-Out resampling strategy \cite{hero2006geometric}, where each holdout sample corresponds to an entire segment.  

Note that all classifiers were learned from a corrupted training set. Since the test set is also corrupted we used an anomaly detection algorithm (GEM with $5\%$ false alarm rate) to produce a test set with few anomalies, called the nominal test set. This allows us to report the performance of the various algorithms on both the clean test data and on the corrupted test data. Table \ref{tab: class_accuracy_norm} shows the classification accuracy of the methods (trained on the training set alone) applied to nominal test set and Table \ref{tab: class_accuracy} shows the result on the entire corrupted test set. 
For ROD only $\rho=0.02$ and $\rho=0.20$ are shown; it was determined that $\rho = 0.20$ achieves the best performance in the range $\rho\in [0.01,1]$. In Table \ref{tab: class_accuracy_norm}, it is seen that the GEM-MED method outperforms the ROD-$\rho$ algorithms for all values of $\rho$ as a function of classification accuracy when individual sensors 1,2,4 are used. Notice that  when used alone neither kernel MED nor kernel SVM is resistant to the sensor failures in the training set, which explains their poor accuracy in sensor 3 and sensor 4. Also in the column \emph{GEM$+$MED} of Table \ref{tab: class_accuracy_norm}, we first trained a GEM anomaly detector to screen out  $5\%$ of the noisy training set, then trained a MED classifier on the rest of the training data. Note that GEM-MED learns both the detector and the classifier jointly on noisy training data. Table \ref{tab: class_accuracy_norm} shows that the two stage training approach has poor performance in highly corrupted sensors 3 and 4. This is due to the fact that when the GEM detector is learned without inferring the classification margin, it cannot effectively limit the negative influence of those corrupted samples that are close to the class boundary.  In Table \ref{tab: class_accuracy}, we show the classification accuracy when both the nominal and anomalous test samples are involved in evaluation. We observe a performance degradation for all methods due to the irregularity of the outliers in the test set. In spite of this, the GEM-MED maintains a superior performance over all other methods. 
This reflects the superiority of the proposed joint classification and detection approach of GEM-MED as compared with \emph{GEM $+$ MED} approach.   

\begin{table}[tb]
\centering
\renewcommand{\arraystretch}{2.0}
 \caption{\footnotesize Anomaly detection accuracy with different sensors, with the best performance shown in \textbf{bold}.}
\label{tab: detect_accuracy} 
\begin{tabularx}{240pt}{|m{40pt}<{\centering}|m{50pt}<{\centering}|m{50pt}<{\centering}|m{50pt}<{\centering}|}
\hline 
\multicolumn{4}{|c|}{Anomaly Detection Accuracy ($\%$) mean $\pm$ standard error} \\ \hline
sensor no.&  ROD-$0.02$ & ROD-$0.2$ & GEM-MED \\ 
\hline 
$1$ & $30.2 \pm 1.3$ & $59.0 \pm 3.5$ & $\mathbf{70.5 \pm 1.3}$ \\ 
\hline 
$2$ & $23.5 \pm 2.6$ & $\mathbf{63.5 \pm 2.8}$ & $63.4 \pm 2.5$ \\ 
\hline 
$3$ & $5.3 \pm 1.4$ & $48.1\pm 3.3$ & $\mathbf{72.8 \pm 1.5}$ \\ 
\hline 
$4$ & $22.8 \pm 3.2$ & $65.2 \pm 4.2$ & $\mathbf{88.1 \pm 2.1}$ \\ 
\hline
$1,2,3,4$ & $38.5 \pm 6.3$ & $63.3 \pm 5.5$ & $\mathbf{88.5 \pm 4.1}$ \\ 
\hline 
\end{tabularx}  \vspace{3pt}
\end{table}

Table \ref{tab: detect_accuracy} compares the anomaly detection accuracies on both \emph{training and test data} for ROD and GEM-MED, where the accuracy is computed relative to ground truth anomalies.  Note that GEM-MED has  significant improvement in accuracy over ROD when trained individually  on sensors 1,3,4, respectively, and when trained  on all of the combined sensors. When trained on sensor 2 alone, the accuracies of GEM-MED and ROD-0.2 are essentially equivalent. In sensor 2 the anomalies appear to occur in concentrated bursts and we conjecture that that a GEM-MED model that accounts for clustered and dependent anomalies may be able to do better. Such an extension is left to future work.   

\vspace{-20pt}
\section{Conclusion}\label{lab: conclude}
In this paper we proposed a unified GEM-MED approach for anomaly-resistant classification. We demonstrated its performance advantages in terms of both classification accuracy and detection rate on a simulated data set and on a real footstep data set, as compared to an anomaly-blind Ramp-Loss-based classification method (ROD). Further work could include generalization to the setting of multiple sensor types where anomalies exist in both training and test sets. \vspace{-10pt} 

\section{Acknowledgment}\vspace{-10pt}
This work was supported in part by the U.S. Army Research Lab under ARO grant WA11NF-11-1-103A1. 
We also thanks Xu LinLi and Kumar Sricharan for their inputs on this work.\vspace{-10pt}
\appendix
\section{Appendix}\vspace{-10pt}
\subsection{Derivation of theorem \ref{prop: unique_solution}}\label{app: prop_1}\vspace{-10pt}
\begin{proof}
The proof of the convexity of the problem can be seen in chapter $12$ of the standard textbook \cite{jaakkola1999maximum}, since the problem is with respect to the distribution $q$. The uniqueness of the solution follows directly from the fact that the problem is convex.  

The Lagrangian function is given as
\begin{align*}
&\cL(q, \mb{\lambda}, \mb{\mu}, \mb{\nu}) \\
&=  \E{q}{\log q - \log p_{0}} + \sum_{n\in T}\lambda_{n}\E{q}{\eta_{n}\cL_{C}}- \sum_{z\in \set{\pm 1}}\mu_{z} \E{q}{\tilde{\cL}_{D,z}}\\
&\phantom{=} -\sum_{z\in \set{\pm 1}}\kappa_{z}\E{q}{\sum_{n: y_{n}=z}\eta_{n}/\abs{T}- \hat{\beta}}
\end{align*} with dual variables $\mb{\lambda}=\set{\lambda_{n}, n\in T} \succeq \mb{0}$, $\mb{\mu}= (\mu_{z}, z\in \pm 1)\succeq \mb{0}$ and $\nu\ge 0$.

Then the result follows directly from solving a system of equations according to the KKT condition. 
\end{proof}\vspace{-10pt}
\subsection{Derivation of theorem \ref{prop: dual_opt}}\label{app: prop_2}\vspace{-10pt}
\begin{proof}
According to \cite{jaakkola1999maximum},  the dual optimization is given as 
\begin{align*}
&\max_{\mb{\lambda}, \mb{\mu}, \mb{\kappa} \ge 0}  -\log Z(\mb{\lambda}, \mb{\mu}, \mb{\kappa})\\
&=-\log   \prod_{n\in T}\int \exp\paren{ -c(1-\xi_{n})- \lambda_{n}\xi_{n}}d\xi_{n}  \\
&\phantom{=}\times \int\int \exp\paren{-\frac{1}{2}\mb{f}^{T}\mb{K}^{-1}\mb{f}+ \sum_{n}\lambda_{n}\eta_{n}y_{n}f_{n}}d\mb{f} \\
& \phantom{=}\times p_{0}(\mb{\eta})  \exp\left(-\sum_{z\in \set{\pm 1}}\mu_{z}\sum_{n: z}\eta_{n}d_{n}+\sum_{z\in \set{\pm 1}}\mu_{z}\hat{\gamma}_{z} \right.\\
& \phantom{===========} \left. + \sum_{z\in \set{\pm 1}}\kappa_{z}\sum_{n: z}\eta_{n} + \sum_{z\in \set{\pm 1}}\kappa_{z}\hat{\beta} \right)   d\mb{\eta} \\
&= \sum_{n\in T}\paren{\lambda_{n}+ \log\paren{1- \lambda_{n}/c}}-\sum_{z\in \set{\pm 1}}\mu_{z}\hat{\gamma}_{z}-  (\sum\kappa_{z})\hat{\beta} \\
&\phantom{=}  -\log \int \exp\paren{ \frac{1}{2}Q(\mb{K}, \paren{\mb{\lambda}\odot \mb{\eta} \odot \mb{y} })+ \mb{\eta}^{T}\paren{-\mb{\mu}\otimes \mb{d} + \mb{\kappa}\otimes \mb{e}}   }\\
&\phantom{==========} \times p_{0}(\mb{\eta})d\mb{\eta}  
\end{align*}
where
\begin{align*}
Q(\mb{K}, \mb{x}) &= \mb{x}^{T}\mb{K}\mb{x}\\
Q(\mb{K}, \paren{\mb{\lambda}\odot \mb{\eta}}) &\defeq \paren{\mb{\lambda}\odot \mb{\eta}}^{T}\mb{K}\paren{\mb{\lambda}\odot \mb{\eta}}\\
&= \mb{\lambda}^{T}\paren{\mb{K}\odot (\mb{\eta}\mb{\eta}^{T})}\mb{\lambda}\\
&= Q(\mb{K}(\mb{\eta}), \mb{\lambda}).
\end{align*}
\end{proof}\vspace{-10pt}
\subsection{Derivation of \eqref{eqn: post_eta}, \eqref{eqn: post_f}}\label{app: lemma_1}\vspace{-10pt}
\begin{proof}
The expression for $q(\overline{\Theta})$ is given as
\begin{align*}
&q(\overline{\Theta})\propto \exp\paren{-\frac{1}{2}\mb{f}^{T}\mb{K}^{-1}\mb{f}+ \sum_{n}\lambda_{n}\eta_{n}y_{n}f_{n}} \\
& \phantom{=}\times p_{0}(\mb{\eta})  \exp\left(-\sum_{z\in \set{\pm 1}}\mu_{z}\sum_{n: z}\eta_{n}d_{n} + \sum_{z\in \set{\pm 1}}\kappa_{z}\sum_{n: z}\eta_{n} \right)\\
&\phantom{=} \times \prod_{n\in T}\exp\paren{-c+ (c-\lambda_{n})\xi_{n}}\\
&=q(f, \mb{\eta})\prod_{n}q(\xi_{n})
\end{align*}
Given all $\eta_{n}, n\in T$, 
\begin{align*}
&q(f| \mb{\eta})\propto \exp\paren{-\frac{1}{2}\mb{f}^{T}\mb{K}^{-1}\mb{f}+ \sum_{n}(\lambda_{n}\eta_{n})f_{n}}\\
&= \exp\paren{ -\frac{1}{2}\paren{ \mb{f} - \mb{K}\paren{\mb{\lambda}\odot\mb{\eta}\odot \mb{y}} }^{T}\mb{K}^{-1}\paren{ \mb{f} - \mb{K}\paren{\mb{\lambda}\odot \mb{\eta}\odot \mb{y}} }}\\
&= \cN(\mb{K}\paren{\mb{\lambda}\odot \mb{\eta}\odot  \mb{y}}, \mb{K}).
\end{align*}
On the other hand, given $f$, $\mb{\eta} = (\eta_{n}, n\in T)$ are fully separated in above formula, therefore $q(\mb{\eta}|f) = \prod_{n}q(\eta_{n}|f)$. 
\end{proof}
\subsection{Implementation of Gibbs sampler}\label{app: Gibbs}\vspace{-10pt}
We implement a Gibbs sampler \cite{robert2013monte} to estimate $\E{q(f, \boldsymbol{\eta})}{G(f, \boldsymbol{\eta})}$, where $G$ is a general function of $f$ and $\boldsymbol{\eta}$, as expressed in  \eqref{eqn: grad_lambda}, \eqref{eqn: grad_mu}, \eqref{eqn: grad_kappa}.  The following procedure is applied iteratively \vspace{-10pt}
\begin{itemize}
\item Initialization: Set $\hat{\boldsymbol{\eta}}_{0} = [1,\ldots, 1]^{T}$ and set a fixed dual parameter $( \boldsymbol{\lambda}, \boldsymbol{\mu},\boldsymbol{\kappa})$. Let $G_{0}= 0$.
\item For each $t=1,2,\ldots, T_{G}$ or until convergence
\begin{enumerate}
\item Given $\hat{\mb{\eta}}_{t-1}= (\hat{\eta}_{n,t-1})$,  generate decision value $f_{t}(\mathbf{x}_{n}), n=1,\ldots,N$ according to the Gaussian process \eqref{eqn: postfix} with mean function $\hat{f}_{t}(\cdot) = \sum_{n\in T}\lambda_{n}\hat{\eta}_{n, t-1}y_{n}K(\cdot, \mb{x}_{n})$.
\item Given $\set{f_{t}(\mathbf{x}_{n})}_{1\le n\le N}$, for $r=1,\ldots, N_{r}$,
\begin{enumerate}
\item generate latent variables $\eta_{n,t}^{(r)}\in \set{0,1}$ according to the Bernoulli distribution with parameter as in \eqref{eqn: post_eta} for each $n$ independently. 
\end{enumerate} 
\item Compute the sample mean of $\hat{\eta}_{n,t} = \frac{1}{N_{r}}\sum_{r=1}^{N_{r}}\eta_{n,t}^{(r)}$ $ \in [0,1], n=1,\ldots, N$. Let $\hat{\boldsymbol{\eta}}_{t} = (\hat{\eta}_{n,t} )_{1\le n\le N}$.
\item Evaluate $G_{t}$ via $G_{t} = \frac{t-1}{t}G_{t-1} + \frac{1}{t}G(\hat{f}_{t},\hat{\boldsymbol{\eta}}_{t})$
\end{enumerate}
\item Output the approximate expectation $\hat{\mathds{E}}_{q(f, \boldsymbol{\eta})}\brac{G(f, \boldsymbol{\eta})}= G_{T}$ as well as the mean  estimate $\hat{\boldsymbol{\eta}}_{T}$ and $\hat{f}_{T}(\mb{x}_{n}), 1\le n\le N$ when the Gibbs chain process becomes stationary. \\[-30pt]
\end{itemize}

\bibliographystyle{IEEEtran}
\bibliography{tianpei_Journal_May_2014.bib}
\end{document}